\documentclass[10pt,journal,compsoc]{IEEEtran}

\usepackage[table]{xcolor}
\usepackage{graphicx}
\usepackage{soul}
\usepackage{url}
\usepackage{hyperref}
\usepackage{verbatim}
\usepackage{subfigure}
\usepackage{calc}
\usepackage{amstext,amsmath,amssymb,bm}
\usepackage{stackengine}
\newcommand\IncG[2][]{\addstackgap{%
\raisebox{-.5\height}{\includegraphics[#1]{#2}}}}

\usepackage{xcolor}
\usepackage{paralist}
\usepackage{multirow}
\usepackage{chngpage}
\usepackage{lipsum}
\usepackage{dutchcal}

\usepackage{makecell}
\usepackage{algorithm}  
\usepackage{algpseudocode}

\ifCLASSOPTIONcompsoc
  \usepackage[nocompress]{cite}
\else
  \usepackage{cite}
\fi
\ifCLASSINFOpdf
\else
\fi
\begin{document}
%
\title{Deep Depth Completion from Extremely Sparse Data: A Survey}
%
%
%
%

\author{Junjie Hu,~\IEEEmembership{Member,~IEEE,}
Chenyu Bao,
Mete Ozay,
Chenyou Fan,
Qing Gao,
Honghai Liu,~\IEEEmembership{Fellow,~IEEE}
and Tin Lun Lam,~\IEEEmembership{Senior Member,~IEEE}
\IEEEcompsocitemizethanks{\IEEEcompsocthanksitem J.Hu, Q.Gao are with the Shenzhen Institute of Artificial Intelligence and Robotics for Society (AIRS), China.
E-mail: hujunjie@cuhk.edu.cn,  gaoqing@cuhk.edu.cn.
\IEEEcompsocthanksitem M.Ozay is with the Samsung Research, UK.  E-mail: meteozay@gmail.com.
\IEEEcompsocthanksitem C.Fan is with the School of Artificial Intelligence, South China Normal University, China. E-mail: fanchenyou@scnu.edu.cn.
\IEEEcompsocthanksitem H.Liu is with the State Key Laboratory of
Robotics and systems, Harbin Institute of Technology (Shenzhen), China. E-mail: honghai.liu@hit.edu.cn.
\IEEEcompsocthanksitem C.Bao and T.T.Lam are with the Chinese University of Hong Kong, Shenzhen, China.
E-mail: boggy615@gmail.com, tllam@cuhk.edu.cn.
\IEEEcompsocthanksitem J.Hu and C.Bao contribute equally.
\IEEEcompsocthanksitem T.L.Lam is the corresponding author.}
}

%
%

\markboth{Journal of \LaTeX\ Class Files,~Vol.~14, No.~8, May~2022}%
{Shell \MakeLowercase{\textit{et al.}}: Bare Demo of IEEEtran.cls for Computer Society Journals}
%



\IEEEtitleabstractindextext{%
\begin{abstract}

Depth completion aims at predicting dense pixel-wise depth from an extremely sparse map captured from a depth sensor, e.g., LiDARs. It plays an essential role in various applications such as autonomous driving, 3D reconstruction, augmented reality, and robot navigation. Recent successes on the task have been demonstrated and dominated by deep learning based solutions. In this article, for the first time, we provide a comprehensive literature review that helps readers better grasp the research trends and clearly understand the current advances. We investigate the related studies from the design aspects of network architectures, loss functions, benchmark datasets, and learning strategies with a proposal of a novel taxonomy that categorizes existing methods. Besides, we present a quantitative comparison of model performance on three widely used benchmarks, including indoor and outdoor datasets. Finally, we discuss the challenges of prior works and provide readers with some insights for future research directions.
\end{abstract}

}

\maketitle

\IEEEdisplaynontitleabstractindextext

%
\IEEEpeerreviewmaketitle

\IEEEraisesectionheading{\section{Introduction}\label{sec:introduction}}

\IEEEPARstart{A}CQUIRING correct pixel-wise scene depth plays a substantial role in various tasks such as scene understanding \cite {max-S-and-D}, autonomous driving \cite{song2021self}, robotic navigation \cite{ma2019sparse,aerial_depth}, simultaneous localization and mapping \cite{XiyueGuo2021}, intelligent farming \cite{farkhani2019sparse}, and augmented reality \cite{Du2020DepthLab}. Thus, it has been a long-term goal studied in past decades. 
One cost-effective way of obtaining scene depth is to directly estimate it from a single image with monocular depth estimation algorithms \cite{Godard2017UnsupervisedMD,Laina2016DeeperDP,Hu2019RevisitingSI,Fu2018DeepOR}. However, visual methods often yield a low inference accuracy and poor generalizability and thus are vulnerable to real-world deployment.

\begin{figure}[t]
    \centering
    \includegraphics[width = 0.98\linewidth]{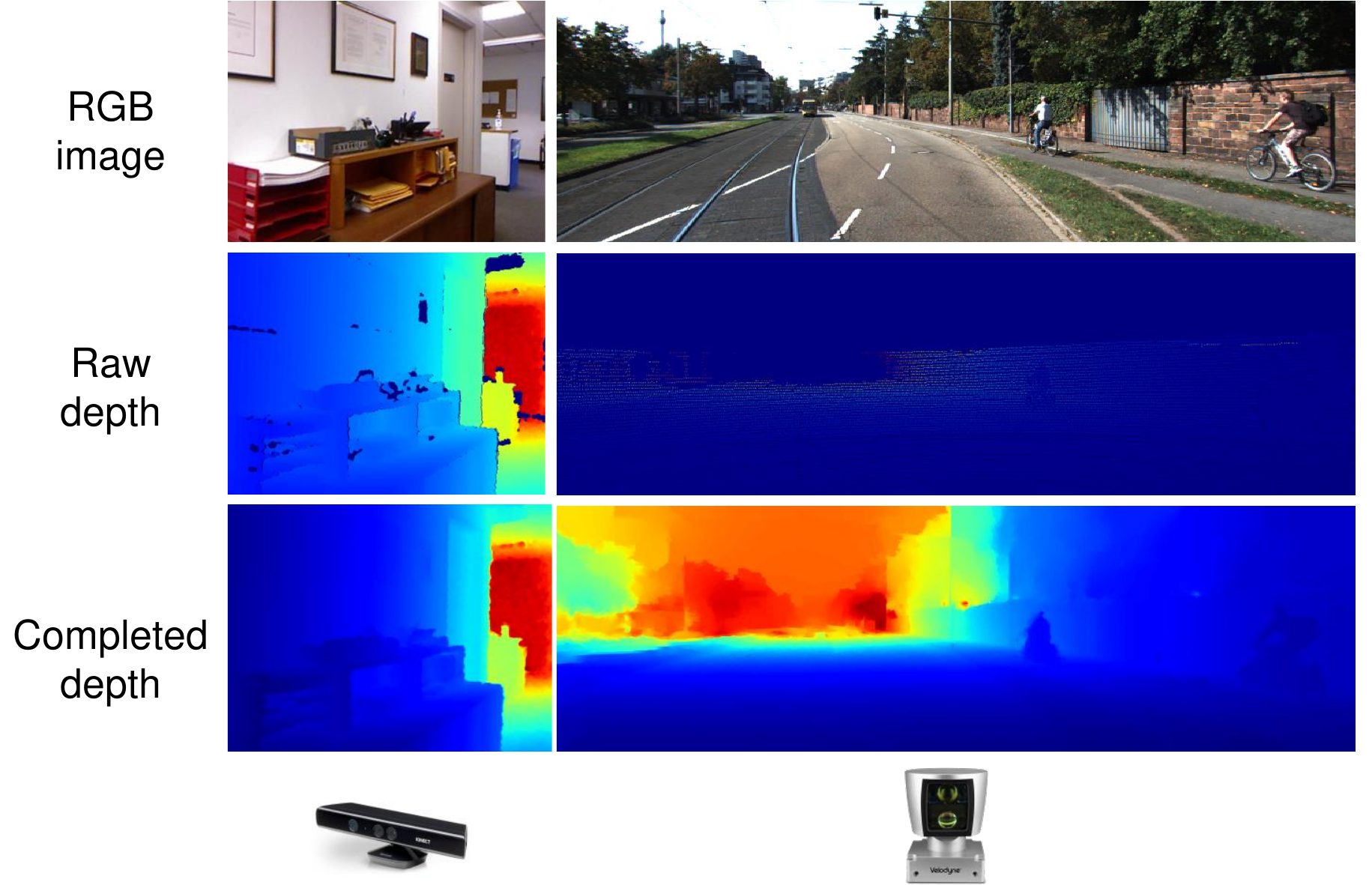}
    \vspace{-2mm}
    \caption{Comparison between captured depth maps by different sensors. The raw sparse depth maps are shown in the middle. The left one is captured by a Kinect in an indoor scenario, and the right one is captured by a LiDAR in an outdoor street. Clearly, the map captured by LiDAR is significantly more sparse. The bottom row shows the completed depth map from the raw sparse map. 
    }
    \label{fig-KITTI-RAW}
\end{figure}

On the other hand, depth sensors provide accurate and robust distance measurements with true scene scales. Therefore, they are more applicable for applications that require a security guarantee and high performance \cite{S-d-selfsuper,song2021self,fu2019lidar}, e.g., self-driving cars. In fact, measuring depths with LiDARs is probably still the most deployable way to obtain reliable depth in industrial applications.
 However, neither LiDAR nor commonly used RGBD cameras, like Microsoft Kinect, can provide a dense pixel-wise depth map. As shown in Fig. \ref{fig-KITTI-RAW}, the depth map captured by Kinect has small holes and the map captured by LiDAR is significantly more sparse. 
It is, therefore, necessary to fill the void pixels in practice.

Since there is a clear difference among depth maps captured by different sensors,
the completion problem and solution are usually sensor-dependent. For example, it is frequently called depth enhancement \cite{shen2013layer,lu2014depth,huang2019hms}, depth inpainting \cite{miao2012texture,liu2012guided} and depth denoising \cite{fu2012kinect,shen2013layer} in many works, where the goal is to infer missing depth values from dense raw depth maps and eliminate outliers (typically, the density is over 80\% as discussed in \cite{S-d-selfsuper}). In this article, we particularly focus on the completion task for extremely sparse data, e.g., for depth maps captured by LiDARs where the sparsity is usually over 95\%. This problem is studied and handled separately in related literature 
and is much more challenging due to the low density of the sparse input. For simplicity, we refer to depth completion from extremely data as depth completion in the rest of the article.

\begin{figure*}[t!]
\begin{center}
\includegraphics[width = \linewidth]{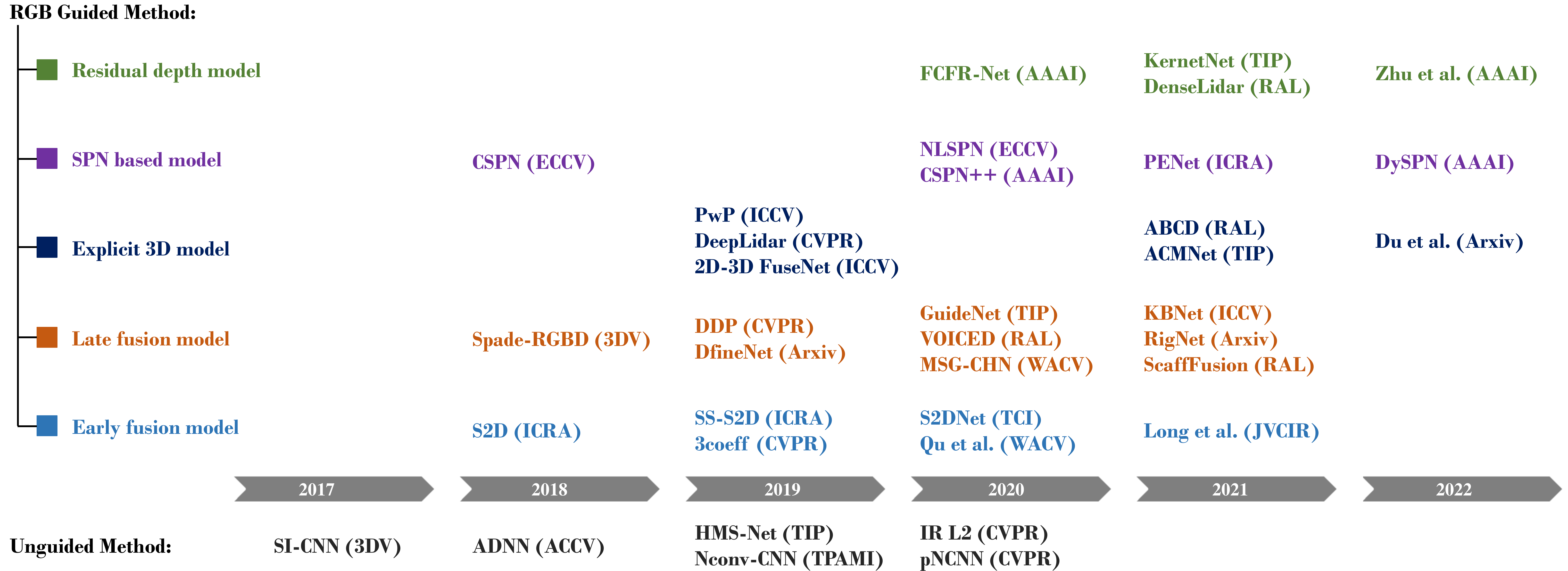}
\vspace{-5mm}
\caption{A timeline for deep learning based depth completion methods. We show some selected works to visualize the evolution process. Unguided methods: SI-CNN \cite{SICNN}, ADNN \cite{chodosh2018deep}, HMS-Net \cite{huang2019hms}, Ncon-CNN \cite{eldesokey2019confidence}, IR L2 \cite{from_depth_what}, pNCNN \cite{eldesokey2020uncertainty}. RGB guided methods: 1) Early fusion models: S2D \cite{S-D-single-image}, SS-S2D \cite{S-d-selfsuper}, 3coeff \cite{depth_coefficient}, S2DNet \cite{hambarde2020s2dnet}, Qu et al. \cite{qu2020depth}, Long et al. \cite{long2021depth}. 2) Late fusion models: Spade-RGBD \cite{max-S-and-D}, DDP \cite{ddp}, DfineNet\cite{zhang2019dfinenet}, GuideNet \cite{learning-guided}, VOICED \cite{wong2020unsupervised}, MSG-CHN \cite{cascade}, KBNet \cite{wong2021unsupervised}, RigNet \cite{yan2021rignet}, ScaffFusion \cite{wong2021scaffnet}. 3) Explicit 3D representation models: PwP \cite{xu2019depth}, DeepLidar \cite{DeepLiDAR}, 2D-3D fuseNet \cite{2d-3d}, ABCD \cite{jeon2021abcd}, ACMNet \cite{zhao2021adaptive}, Du et al. \cite{Du2022DepthCU}. 4) SPN-based models: CSPN \cite{cspn}, NLSPN \cite{nlspn}, CSPN++ \cite{cspn++}, PENet \cite{penet}, DySPN \cite{Lin2022DynamicSP}. 5) Residual depth models: FCFR-Net \cite{fcfr}, KernelNet \cite{liu2021learning}, DenseLiDAR \cite{gu2021denselidar}, Zhu et al. \cite{zhu2021robust}. }
 \label{fig-history}    
\end{center}
\end{figure*}

In recent years, deep learning based methods have shown compelling performance on the task and have led the development trend. 
It is shown in prior works that a network with several convolutional layers \cite{SICNN}, or a simple auto-encoder\cite{Lu2022DepthCA} can complete missing depths.
Moreover, depth completion can be further improved by leveraging RGB information.
A typical method of this type \cite{max-S-and-D,shivakumar2019dfusenet} is to use dual encoders for extracting features from a sparse depth map and its corresponding RGB image, respectively, and later fuse them with a decoder. 

To push the envelope of depth completion, recent approaches tend to use complicated network structures and complex learning strategies. In addition to multi-branches used for feature extraction from multi-modality data, e.g., image and sparse depths, researchers have begun to integrate surface normal \cite{DeepLiDAR}, affinity matrix \cite{cspn}, residual depth map \cite{gu2021denselidar}, etc., into their frameworks. Besides, to cope with the lack of supervised pixels, some works introduced exploiting multi-view geometric constraints \cite{S-d-selfsuper} and adversarial regularization \cite{khan2021sparse}. These efforts have greatly facilitated the progress in the depth completion task.

Despite the tremendous progress made by learning based approaches, to the best of our knowledge, a comprehensive survey is lacking. 
This article aims to depict the development of learning based depth completion through hierarchically analyzing and categorizing existing methods and provide readers with a straightforward understanding of deep depth completion with some valuable instructions. 
 Typically, we hope to answer the following questions: 
 \begin{enumerate}
     \item What are the common characteristics of previous methods for achieving highly accurate depth completion? 
     
     \item What are the pros and cons of RGB guided approaches compared to unguided methods? 
     
     \item Since most previous works employed both visual and LiDAR data, what are the most effective strategies for multi-modal data fusion? 
     
     \item What are the current challenges?
  \end{enumerate}
 
With the above questions being considered, we survey the related works from January 2017 to May 2022 (at the time of writing). Fig.~\ref{fig-history} visualizes the timeline of the selected methods based on the proposed taxonomy, where the bottom and the top show the unguided and five types of RGB guided methods, respectively. It is seen that although early studies tackle depth completion in an unguided fashion, we observed that studies published after 2020 have been gradually dominated by RGB guided methods.
In this article, we investigate the previous studies from the aspects of network structure, loss function, learning strategy, and benchmark datasets. 
We especially stress methods with the proposal of novel algorithms or significant performance boosts and properly provide visual descriptions of their technical contributions to promote the clarification.
Furthermore, we provide quantitative comparisons of existing methods with essential characteristics on the most popular benchmark datasets.
Through the in-depth analysis of previous studies, we wish the reader can gain a clear understanding of deep depth completion.

\begin{table*}[t]
\caption{A brief overview of the proposed taxonomy.}
\renewcommand\arraystretch{1.2}
\begin{center}
\begin{tabular}
{|m{0.18\textwidth}<{\centering}|m{0.35\textwidth}<{\raggedright}|m{0.38\textwidth}<{\raggedright}|}
\hline
Main categories & \multicolumn{1}{c|}{Sub-categories} & \multicolumn{1}{c|}{Major characteristics}\\ \hline
\multirow{6}{*}{\makecell{Unguided methods \\ (Sec.~\ref{sec-unguided})}} 
&Sparsity-aware CNNs (\textbf{SACNN}, Sec.~\ref{sparsity-aware-cnn}) & Using the binary validity mask to indicate missing elements during convolution.\\ \cline{2-3}
&Normalized CNNs (\textbf{NCNN}, Sec.~\ref{normalized-cnn}) & 1). Built on normalized convolution  2). Replacing the validity mask with continuous confidence mask. \\ \cline{2-3}
&Training with Auxiliary Images (\textbf{TwAI}, Sec.~\ref{auxiliary image learning})& 
Integrating image reconstruction into latent or output space to encourage learning semantic cues. 
Image guided training and unguided inference are employed.\\ 
\hline
\multirow{16}{*}{\makecell{RGB guided methods \\ (Sec.~\ref{sec-guided})}} 
& Early fusion models (\textbf{EFM}, Sec.~\ref{early-fusion})  
\begin{itemize}
\item Encoder-decoder networks (\textbf{EDN}, Sec.~\ref{encoder-decoder-network})  
\item Coarse to refinement prediction (\textbf{C2RP}, Sec.~\ref{C2RP}) 
 \vspace{-2.5mm}
\end{itemize}

 & Directly aggregating the image and sparse depth map input or fusing the multi-modality features at the first convolutional layer.\\ \cline{2-3}
&  

Late fusion models (\textbf{LFM}, Sec.~\ref{late-fusion}) 
\begin{itemize}
\item Dual-encoder networks (\textbf{DEN}, Sec.~\ref{DEN}) 
\item Double encoder-decoder networks (\textbf{DEDN}, Sec.~\ref{DEDN}) 
\item Global and Local Depth Prediction (\textbf{GLDP}, Sec.~\ref{GLDPN})  
 \vspace{-2.5mm}
\end{itemize}

& The framework usually consists of dual encoders or two sub-networks; the one is used for extracting RGB features and the other is used for extracting depth features. Fusion is conducted at the intermediate layers, e.g., fusing extracted features from encoders. \\ \cline{2-3}
& Explicit 3D representation models (\textbf{E3DR}, Sec.~\ref{E3DR})
\begin{itemize}
\item 3D-aware convolution (\textbf{3DAC}, Sec.~\ref{3d-awae}) 
\item Intermediate surface normal representation (\textbf{ISNR}, Sec.~\ref{surface-normal}) 
\item Learning from point clouds (\textbf{LfPC}, Sec.~\ref{point-clouds})  
 \vspace{-2.5mm}
\end{itemize}

&Explicitly learning 3D representations, such as applying 3D convolutions, embedding surface normals, and learning from 3D point clouds.\\ \cline{2-3}
& Residual depth models (\textbf{RDM}, Sec.~\ref{RDM}) & Learning a coarse depth map and a residual depth map. Their combination generates the final depth map. \\ \cline{2-3}
& SPN-based models (\textbf{SPM}, Sec.~\ref{SPNM})  & 1). Based on the spatial propagation network. 2). First learning the affinity matrix, and then applying affinity based depth refinement.\\ 
\hline
\end{tabular}
\end{center}
\label{table-taxonomy}
\end{table*}

In summary, our key contributions are as follows:
\begin{itemize}
    \item To the best of our knowledge, this is the first survey for depth completion. We give an in-depth and comprehensive review, including both unguided and RGB guided methods. 
    \item We propose a novel taxonomy to categorize previous methods and visualize their main characteristics, including network structures, loss functions, and learning strategies.
    \item The article covers the most advanced and recent progress of deep learning based depth completion with performance comparison on benchmark datasets. It provides readers with state-of-the-art methods.
    \item  We summarize the pros and cons of methods in each type via analyzing their accuracy and model complexity. 
    \item We provide several open issues and promising future research directions.
\end{itemize}

The remainder of this article is organized as follows: Section \ref{premilinary} gives the formulation of deep learning based depth completion and provides the proposed taxonomy. Section \ref{sec-unguided} reviews unguided methods, and Section \ref{sec-guided} elaborates RGB guided methods.
Section \ref{sec-loss} introduces the loss functions employed in previous approaches. Section \ref{sec-datasets} lists the benchmark datasets and introduces the evaluation metrics for the depth completion task.
Section \ref{result} compares the previous methods from comprehensively different perspectives.
 Section \ref{discussion} summarizes the open challenges and provides valuable directions for future research. Section \ref{conclusion} gives the conclusion.

\section{Deep Learning Based Depth Completion}
\label{premilinary}
In this section, we first give a common formulation of the depth completion task. Then, we outline the proposed taxonomy. 
Noting that some methods share common characteristics, we group them by jointly considering network structures and main technical contributions. 

\subsection{Problem Formulation}
In depth completion, a deep neural network $N$ with parameters $\mathcal{W}$ predicts a dense depth map $\hat{Y} \in \hat{\mathcal{Y}}$ of a given sparse depth map $Y' \in \mathcal{Y'}$  by
\begin{equation}
    \hat{Y} = N\left (Y';\mathcal{W} \right).
    \label{eq_dc}
\end{equation}

\textbf{Unguided depth completion:} In \eqref{eq_dc}, depth completion is performed using only the sparse input without guidance from different modality data. Therefore, it is called unguided depth completion. These methods are reviewed in detail in Section~\ref{sec-unguided}.

\textbf{RGB guided depth completion:} In many works, both the sparse depth map and its corresponding RGB image are utilized for inputs. In this case, the task is formulated by
\begin{equation}
     \hat{Y} = N\left (Y', I;\mathcal{W}  \right )
    \label{eq_dc2}
\end{equation}
where $I$ denotes the RGB image whose pixels are aligned with $Y'$. Then, task employed by \eqref{eq_dc2} is referred to as RGB guided depth completion which is elucidated in Section~\ref{sec-guided}.

The parameters $\mathcal{W}$ of the network $N$ are optimized to train the network by solving
\begin{equation}
    \mathcal{\hat{W}} = \mathop{\mathrm{argmin}}\limits_{\mathcal{W}}\mathcal{L}\left (\hat{\mathcal{Y}}, \mathcal{Y};\mathcal{W} \right )
    \label{eq-loss}
\end{equation}
where $\mathcal{Y}$ denotes the set of ground truth depth maps, and $\mathcal{L}$ is a loss function which is usually defined to penalize pixel-wise discrepancy between the prediction and the ground truth on the valid pixels through back-propagation while training $N$. Depending on the specific learning strategies, some other losses, such as unsupervised photometric loss, adversarial loss, and regularization terms on depth maps, are properly applied. An in-depth discussion of learning objectives and loss functions is given in Section~\ref{sec-loss}.

\subsection{Taxonomy}

In this article, we propose a detailed taxonomy by jointly considering network structures and main technical contributions.
An existing method is firstly categorized into either an unguided method or an RGB guided approach. Then, it is further classified into a more specific sub-category. Table~\ref{table-taxonomy} gives an overview of the proposed taxonomy with descriptions of the major factors for identifying categories.

As seen, unguided methods have three sub-categories, including methods 1) employing sparsity-aware CNNs, 2) employing normalized CNNs, and 3) training with Auxiliary Images.
Guided methods include five sub-categories. Some of them also have more concrete classes.
For the first and second categories, i.e., early fusion and late fusion models, the fusion strategy is the main factor considered in our taxonomy.
For the late three categories, i.e, explicit 3D representation models, residual depth models, and spatial propagation network (SPN) based models, the fusion strategy is not the major factor in identifying their types since they hold distinct characteristics and both early fusion and late fusion are used in previous methods. 

 For methods of each category, we also discuss their pros and cons in the related section. For most methods, we find that their advantage in accuracy is a disadvantage in model complexity, and vice versa.
 Fortunately, most methods provide quantitative results on the standard benchmark datasets. These studies allow us to analyze and compare their performance fairly.

\section{Unguided Depth Completion}

\label{sec-unguided}
Given a sparse depth map, unguided methods aim at directly completing it with a deep neural network model. Previous methods can be generally categorized into three groups: methods using 1) sparsity-aware CNN, 2) normalized CNN, and 3) training with auxiliary images.

\subsection{Sparsity-Aware CNNs}
\label{sparsity-aware-cnn}

\textbf{Overall insight:} \textit{Identifying valid and missing elements with a binary mask during convolution operation enables standard CNNs to perform better for sparse depth inputs. }

Uhrig et al. \cite{SICNN} proposed the first deep learning based unguided method.
They first verified that normal convolutions are not able to handle sparse input as they typically cause mosaic effects and proposed a new sparse convolution operation. Then, they introduced a 6-layers CNN assembled with the proposed sparse convolution.
The sparse convolution uses a binary validity mask to distinguish between valid and missing values and performs convolution among only valid data. The value of the validity mask is determined by its local neighbors via max-pooling.
This first deep learning based method outperforms non-learning methods and shows the potential of deep learning on the task. Moreover, it inspired lots of subsequent studies. 

However, the sparse convolution is not suitable to be directly applied to classical encoder-decoder networks, which can fully leverage the multi-scale features. 
 Huang et al.\cite{huang2019hms} introduced three sparsity invariant (SI) operations, including SI upsampling, SI average, and SI concatenation,
and built an encoder-decoder based HSMNet. They also demonstrated an application using RGB inputs by adding a small branch to HSMNet.

 Chodosh et al.\cite{chodosh2018deep} formulated the depth completion as a multi-layer convolutional compressed sensing problem and proposed an end-to-end multi-layer dictionary learning algorithm. It is achieved by applying compressed sensing to the deep component analysis (DeepCA) objective \cite{murdock2018deep} and optimizing by ADMM (alternation direction method of multipliers). The over-complete dictionaries are learned with a few convolutional layers via back-propagation.


\subsection{Normalized CNNs}
\label{normalized-cnn}

\textbf{Overall insight:} \textit{Replacing a binary validity mask with a continuous confidence map leads to better completion performance. }

The sparsity-aware methods require validity masks to identify missing values for performing convolutions. 
As argued in \cite{Eldesokey2018PropagatingCT,max-S-and-D,eldesokey2019confidence},  validity masks can degrade the model performance due to the saturation of the mask at early layers in CNNs. 
To tackle this issue, inspired by normalized convolution \cite{Knutsson1993NormalizedAD}, Eldesokey et al. \cite{Eldesokey2018PropagatingCT} introduced the normalized convolutional neural network (NCNN) that generates continuous uncertainty maps for depth completion.
The essential difference is that features obtained using the NCNN are weighed with continuous uncertainty maps instead of binary validity masks.  In addition, convolution filters are constrained to be non-negative by the SoftPlus function \cite{glorot2011deep} for faster convergence.

Although NCNN still takes a sparse mask as an initial input, it yields a continuous confidence map to indicate useful information across the intermediate layers.
In reality, disturbed measurements exist due to the LiDAR projection errors. The initial sparse confidence input cannot exclude such noisy inputs.
To solve this problem, Eldesokey et al. \cite{eldesokey2020uncertainty} further developed a self-supervised approach to estimate a continuous input confidence map for suppressing the disturbed measurements with a network. 
NCNN is also applied to RGB guided depth completion in \cite{hua2018ANC,eldesokey2019confidence}.

\subsection{Training with Auxiliary Images}
\label{auxiliary image learning}

\textbf{Overall insight:} \textit{RGB information can be smartly and implicitly utilized for unguided depth completion by introducing an auxiliary task of depth for reconstruction.}

To overcome the lack of semantic cues,
Lu et al. \cite{from_depth_what} employed an auxiliary learning branch in their framework. Instead of directly using an image as input, they only take a sparse depth map as input and simultaneously predict a reconstructed image and a dense depth map. The RGB images are only used in the training stage as a learning objective to encourage acquiring more complementary image features. A similar method is also seen in \cite{DenseLivox} where RGB and normal are used for auxiliary training. In \cite{Lu2022DepthCA}, an auto-encoder is employed to generate RGB data in latent space, and then the auto-encoder predicts the final depth from it.  This method is unsupervised and does not use denser depth maps as ground truths, showing inferior performance compared to \cite{from_depth_what}.
Although these methods are RGB guided in training, they aim at performing unguided depth completion in inference. Therefore, we categorize them into unguided methods.

\subsection{Discussion} 
As the early attempts to solve deep depth completion, 
the sparsity-aware method \cite{SICNN} improved accuracy compared to standard convolutions. However, the performance is poor, especially compared to the subsequent works \cite{huang2019hms,eldesokey2020uncertainty,from_depth_what}. 
There are two underlying reasons for their low accuracy. First, validity masks used in this work are not spatially scale invariant and thus can only be applied to networks simply assembled with several convolutional layers, i.e., lacking discriminability due to low model capacity. 
Second, the validity information obtained from masks tends to disappear after a few convolutional layers in the network. 

The extended work \cite{huang2019hms} enables SI (sparsity invariant) upsampling, SI average/summation, and SI concatenation of feature maps, and thus allows using a hierarchical encoder-decoder network to gain a significant accuracy improvement (41.5\% boost in RMSE) compared to \cite{SICNN}.
 {The method \cite{chodosh2018deep} differs from \cite{SICNN,huang2019hms} as it poses depth completion as a dictionary learning problem, and has its advantage in requiring the minimum model parameters.}

Normalized CNNs \cite{Eldesokey2018PropagatingCT,eldesokey2020uncertainty} mainly targeted the validity loss issue caused by using binary validity masks. There is a decent performance improvement (20.8\%) over \cite{SICNN} derived from applying normalized convolutions \cite{Eldesokey2018PropagatingCT} and a substantial boost (40.0\%) by further considering eliminating outliers that existed in input sparse depth maps \cite{eldesokey2020uncertainty}.
 One advantage of the method proposed in \cite{eldesokey2020uncertainty} compared to the HSMNet \cite{huang2019hms} is that it achieves comparable performance using a lightweight network with only 0.67M parameters \footnote{The number of parameters is unclear for \cite{huang2019hms}.}.
 
Methods using auxiliary images subtly bring RGB guidance into unguided methods by introducing an additional depth to RGB generation task. The RGB information is implicitly aggregated into depth completion modules using a sharing encoder. The method \cite{from_depth_what} substantially boosts the accuracy and is the current state-of-the-art for unguided methods. Moreover, such a strategy for using RGB images does not lead to any increase in the number of parameters in the inference stage. However, their network has more complexity than those proposed in \cite{eldesokey2020uncertainty} (11.67M vs 0.67M) since it takes Inception \cite{szegedy2015going} based encoder and chooses to use larger kernel sizes.
Besides, using additional RGB information to guide the model training will degrade the generalization accuracy of unguided methods in real-world use-cases.

\section{RGB Guided Depth Completion}
\label{sec-guided}

Unguided methods usually underperform RGB guided methods and suffer from blurring effects and distortion of object boundaries. Their inferior performance is attributed to insufficient prior information on natural scenes. As studied in \cite{huang2000statistics}, depth
maps of natural scenes can be decomposed into smooth surfaces and sharp discontinuities in between them; the latter forms step edges in depth maps. This structure is a key property of depth maps. However, when depth maps are extremely sparse, prior information such as neighboring objects and sharp edges are significantly missing;
therefore, it is even intractable to recover complete depth maps using CNNs.

\begin{figure}[h]
    \centering
    \includegraphics[width=0.5\linewidth]{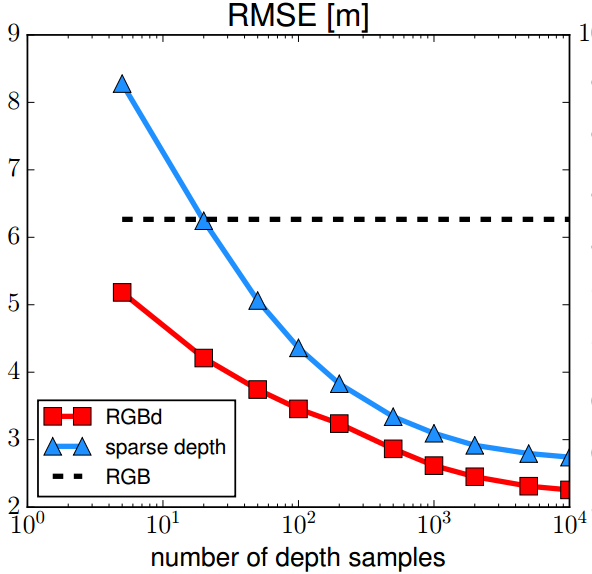}
    \caption{The RMSE for unguided and RGB guided depth completion on the KITTI dataset. From \cite{S-D-single-image}.}
    \label{fig_sparsity}
\end{figure}

Therefore, utilizing RGB information as an additional input is straightforward and reasonable. RGB images provide information about scene structures, including textures, lines and edges, to complement the missing cues of sparse depth maps, and encourage depth continuities inside smooth regions and discontinuities at boundaries. Moreover, they include some monocular cues, e.g., vanishing points \cite{hu2019visualization}, for promoting depth estimation. These benefits complement sparse depth maps.

\begin{figure}[h]
    \centering
    \includegraphics[width=1.0\linewidth]{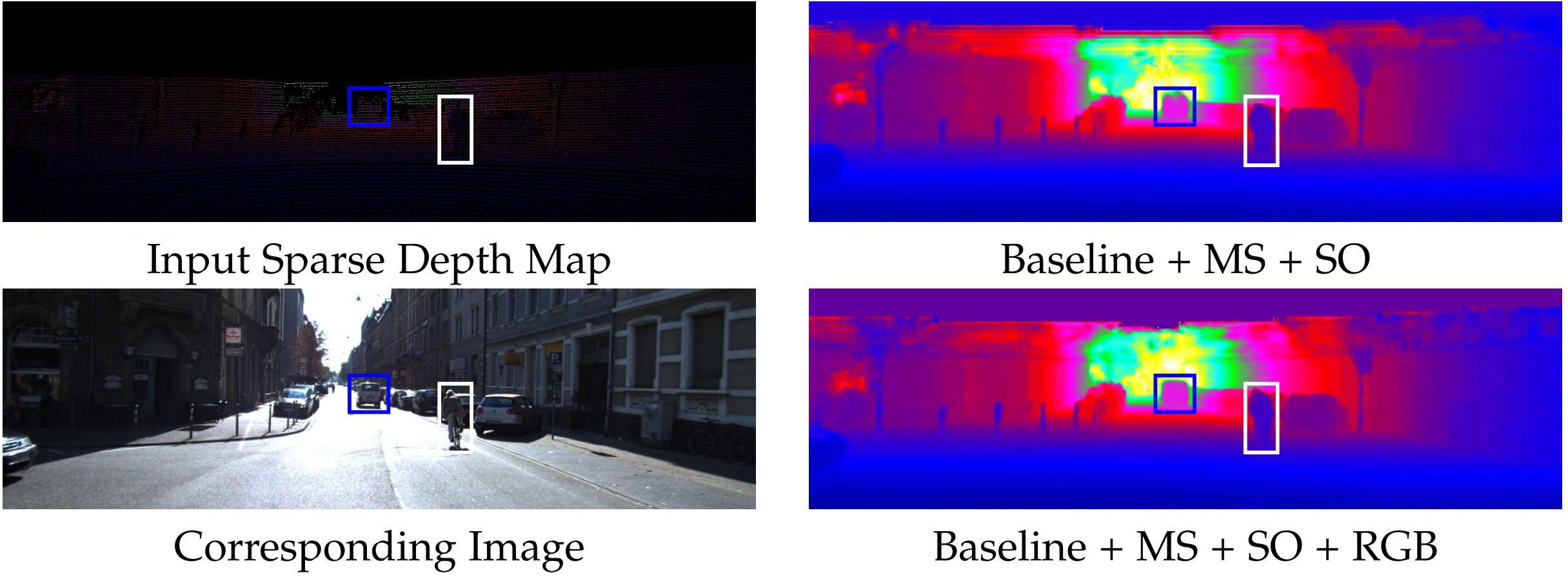}
    \caption{Qualitative comparison of unguided and RGB guided depth completion where MS and SO denote multi-scale structure and sparsity-invariant operations, respectively. From \cite{huang2019hms}.}
    \label{fig_w_wo_img}
\end{figure}

Compared to unguided methods, RGB guided approaches typically have three advantages: i) they generally outperform unguided methods in accuracy, ii) they are more robust to different sparsity levels, and iii) they gain more perceptually correct depth maps. For i) and ii), we can refer to the experimental results shown in Fig.~\ref{fig_sparsity}. As observed, utilizing RGB data improves the accuracy of an unguided model for each sparsity level, and the accuracy degrades slowly when the number of depth samples decreases. For iii), an example of qualitative comparison is given in Fig.~\ref{fig_w_wo_img}. It is shown that RGB guided completion encourages discontinuities at object boundaries while maintaining smoothness inside objects.

To date, different types of methods have been proposed, and they can be categorized into mainly five types: 1) early fusion models, 2) late fusion models, 3) explicit 3D representation models, 4) residual depth models, and 5) spatial propagation network (SPN) based models.

\subsection{Early Fusion Models}
\label{early-fusion}
Early fusion methods directly concatenate a sparse depth map and an RGB image before passing them through a deep model \cite{S-D-single-image,dimitrievski2018learningmorph,DeepLiDAR}, or aggregate multi-modal features at the first convolutional layer of a model \cite{depth_coefficient,xu2019depth,long2021depth}.
Previous methods of early fusion can be divided into two types: methods employing 1) encoder-decoder network and 2) two-stage coarse to refinement prediction. 

\subsubsection{Encoder-decoder Networks}
\label{encoder-decoder-network}

\textbf{Overall insight:} \textit{Early fusion methods built on EDN are straightforward, they are good at model simplicity, yet underperform in accuracy.}

This type of method utilizes a traditional encoder-decoder network (EDN) to solve the pixel-to-pixel regression problem.
An early work is shown in \cite{S-D-single-image} where Ma et al. proposed to accomplish depth completion from both a sparse depth map and its corresponding RGB image. Toward this end, they directly concatenated the RGB image and the sparse depth map and then fed them to an encoder-decoder network built on a ResNet-50 network \cite{He2016DeepRL}. This work also verified that RGB guided depth completion is more accurate and robust than unguided approach for different sparsity levels.

To better enforce the prediction to be consistent with the measurements,
Qu et al. \cite{qu2020depth} replaced the last convolutional layer with a least squares fitting module. In this model, the extracted features obtained from the penultimate layer are treated as a set of bases, and the weights of these bases are obtained through a least squares fit on the depths at valid pixels. As discussed in the paper \cite{qu2020depth}, the method is unable to handle extremely sparse input due to the lack of supervision with enough depth points. 

Motivated by spatially-adaptive denormalization (SPADE) \cite{park2019semantic},
Dmitry et al. \cite{senushkin2020decoder} proposed to learn spatially-dependent scale and bias for normalized features. They introduced a novel decoder assembled with SPADE blocks with a modulation branch. 
The modulation branch takes the valid mask as input and predicts multi-scale modulation signals. These modulation signals are sent to the multiple SPADE blocks in the decoder at each spatial scale to update features. The method's effectiveness has been validated on both indoor depth enhancement and outdoor depth completion.

Instead of the direct concatenation, several approaches \cite{S-d-selfsuper,zhang2019dfinenet,depth_coefficient} used two separate convolutional units to extract features from RGB and depth input at the first layer of the encoder-decoder network, respectively. Then, the multi-modal features were concatenated and sent to the rest of the layers to obtain a complete depth map.

\subsubsection{Coarse to Refinement Prediction}
\label{C2RP}

\textbf{Overall insight:} \textit{Performance of the two-stage coarse to refinement methods highly relies on the quality of pre-estimated depth maps in the first stage of coarse prediction.}

 Some methods employ a two-stage coarse to refinement prediction (C2RP) to achieve more accurate depth estimation.
This kind of methods firstly estimates a coarse depth map in the first coarse prediction stage, then applies the second refinement prediction from the coarse depth map and the RGB image.
For instance, 
Dimitrievski et al. \cite{dimitrievski2018learningmorph} 
 integrated a learnable morphological operator (two contraharmonic mean filter layers \cite{masci2013learning-CHM}) into a U-net \cite{Ronneberger2015UNetCN} based framework. After the morphological operation, the predicted coarse depth map and the RGB image are passed through a U-net to get a refined output.
Similarly, Hambarde et al. \cite{hambarde2020s2dnet} proposed S2DNet which consists of two pyramid networks: S2DCNet and S2DFNet. The S2DCNet performs the first coarse prediction, and the S2DFNet  performs the second refinement.

Unlike the above methods,  several methods proposed to generate multiple maps in the coarse prediction stage. For instance,
Chen et al. \cite{chen2018estimating} generated a dense map with the nearest neighbor interpolation and a prior distance map between depth points based on a Euclidean distance transform of the validity mask. 
The dense map acts as a coarsely predicted map as explored in \cite{dimitrievski2018learningmorph}, and the distance map serves a similar role as the validity mask that informs the model about the valid depth points, but in a different manner from SACNN. As shown in \cite{chen2018estimating}, the inclusion of the distance map improves training stability.
Recently, Hedge et al. \cite{hegde2021deepdnet} proposed the DeepDNet. 
The assumption is that CNNs are considered to learn better features with uniform data rather than randomly distributed data. Therefore, they first convert
 the original sparse input into a grid sparse depth map with quad tree based preprocessing. 
 Then, two coarse maps are generated by applying the nearest neighbor interpolation and Bi-cubic interpolation from the grid sparse map, respectively. 
Such random to uniform transformation gained a slightly better performance than \cite{chen2018estimating} for synthesized depth maps on the NYU-v2 dataset. However, its effectiveness for more realistic scenarios, e.g., KITTI, remains unclear.

In \cite{long2021depth}, depth completion is decomposed into a relative depth estimation and a scale recovery problem. In the first stage, instead of predicting a coarse depth map with a true scale, they propose to estimate a scale-invariant relative depth map from only a single RGB image by isolating the influence of absolute depth values. In the second stage, the relative depth along with the sparse map and RGB image is inputted to a scale prediction network. The final depth map is the multiplication of the relative depth map and its scale map. As argued in \cite{long2021depth}, such design reformulates the completion task in scale space, and thus is more robust for tackling the sparsity.

The idea of rectifying from a coarse prediction is also frequently leveraged in subsequent studies, such as those built on SPNs and residual depth learning frameworks.

\subsubsection{Discussion} 

Early fusion has its advantages in its simplicity, e.g., EDN does not incur many increases in model complexity compared to unguided methods if they are built on the same network.
However, judging from the current situation, early fusion models are rather straightforward as the multi-modal data fusion is simply conducted at the input layer and feature extraction relies entirely on black-box CNNs.
We find that early fusion models generally underperform late fusion models which can learn both domain-specific and correlational features.

 
C2RP is a technical improvement over EDN. It employs an additional encoder-decoder network for prediction refinement at the expense of computational efficiency. For the methods built on C2RP, we find that only S2DNet \cite{hambarde2020s2dnet} and \cite{long2021depth} demonstrate better performance than one-stage prediction methods \cite{depth_coefficient,eldesokey2019confidence,qu2020depth}. 
Although several approaches employing C2RP apply coarse prediction with morphological operator \cite{dimitrievski2018learningmorph}, nearest neighbor interpolation \cite{chen2018estimating,hegde2021deepdnet} and Bi-cubic interpolation \cite{hegde2021deepdnet}, pre-densified depth maps suffer from low quality due to high sparsity of sparse inputs.
Both S2DNet \cite{hambarde2020s2dnet} and \cite{long2021depth} choose to use an encoder-decoder network for coarse prediction and are able to learn more accurate depth maps in the first stage, thus, accordingly improving the final refinement performance.
However, compared to other methods applying refinement, e.g., residual models and SPN models, the refinement lacks spatial constraints.

As these analyses show, the improvements are depicted from the single-stage regression to the two-stage prediction. We find that this tendency generally holds for existing methods that the accuracy improvement largely depends on extending model complexity or sacrificing inference efficiency. 

\subsection{Late Fusion Models}
\label{late-fusion}

Late fusion models usually employ two sub-networks to extract features from (i) RGB images using an RGB encoder network, and (ii) sparse depth inputs using a depth encoder network. The fusion is conducted at intermediate layers of the two sub-networks. 
Most of the previous methods exploit the late fusion strategy with various network structures. Specifically, they are categorized into three types: methods employing 1) dual-encoder network, 2) double encoder-decoder network, and 3) global and local depth prediction.

\subsubsection{Dual-encoder Networks}
\label{DEN}

\textbf{Overall insight:} \textit{Duel-encoder networks (DENs) take a divide-and-conquer strategy that learns domain-specific features from RGB images and sparse depth maps with two separate encoders, respectively. DENs then fuse them to form a correlational feature representation with a decoder.}

Methods built on a dual-encoder network (DEN) commonly use an RGB encoder and a depth encoder for extracting multi-modal features. Then, these features are aggregated and fed into a decoder.
In \cite{max-S-and-D}, Jaritz et al. introduced a two-branch encoder network based on a modified NASNet \cite{nasnet}, where the intermediate features extracted from all encoders are directly concatenated and then outputted to a decoder. 
Notably, Jaritz et al. verified that the validity mask is not necessary for performance improvement for large networks. 
Instead of direct channel-wise concatenation, features extracted from the RGB encoder and the depth encoder are fused in element-wise summation in \cite{shivakumar2019dfusenet,ryu2021scanline}.

Lately, more complicated fusion strategies have been explored.
Fu et al. \cite{fu2020depth} improved the straightforward concatenation of RGB and depth features with an inductive fusion adapted from the conditional neural process \cite{Garnelo2018ConditionalNP}. 
Zhong et al. \cite{zhong2019deep} suggested using the correlation between RGB and depth information. For this purpose,  they proposed the CFCNet which
 extracts the most semantically correlated features  from multi-modal inputs  
 by applying deep canonical
correlation analysis \cite{yang2017canonical} between the sparse depth points and their corresponding pixels in RGB images. 

\begin{figure}[h!]
    \centering
    \includegraphics[width = 0.9\linewidth]{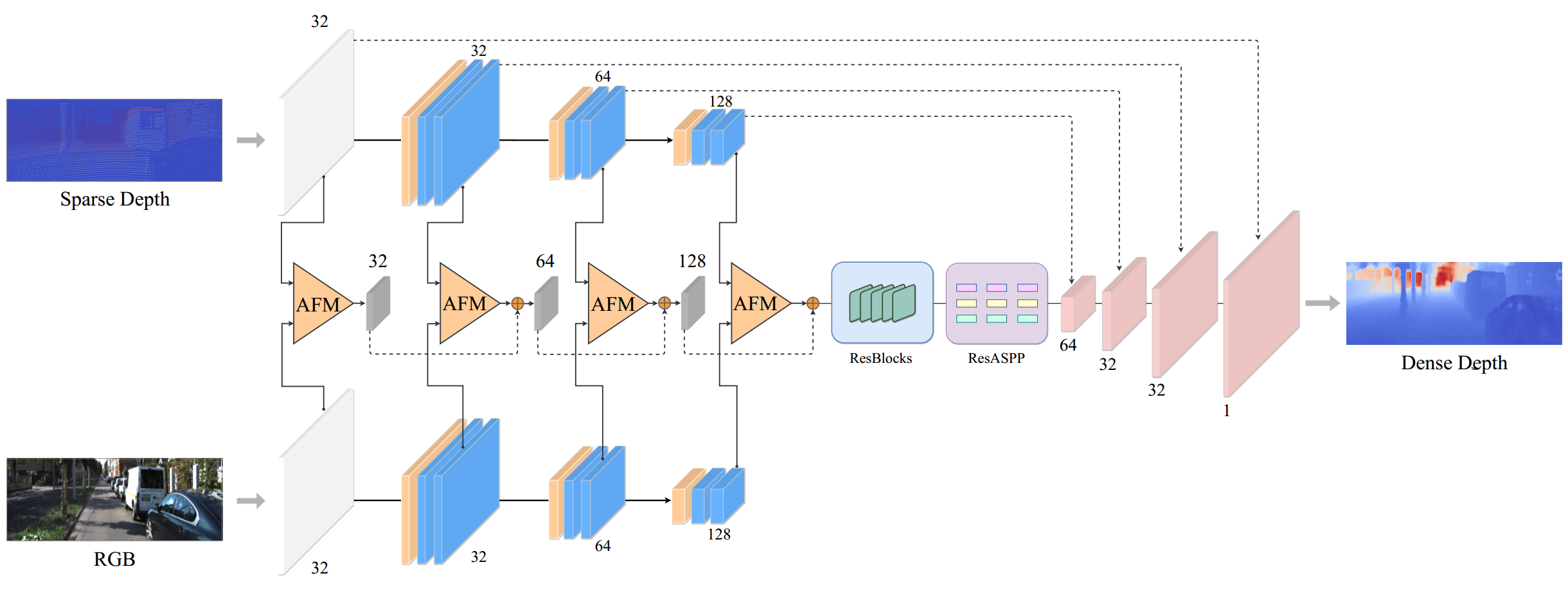}
    \caption{The diagram of the multi-scale adaptation fusion network (MAFN). The framework is a dual-encoder network where features extracted from the RGB encoder and the depth encoder are fused with the adaption fusion (AFM) module at multi-scales. From \cite{zhang2020multiscale}.}
    \label{fig-afm}
\end{figure}

The above approaches only fuse the outputted features from the RGB branch and depth branch at a single spatial scale. They overlook the necessity of fusing RGB and depth features at multiple spatial scales. Multi-scale feature fusion takes advantage of both high-resolution features of shallow layers to avoid structural loss, and low-resolution features of deep layers to improve prediction performance. Several works \cite{mao2016image,zhou2019unet++} have shown that multi-scale feature fusion has an important effect on accuracy in dense prediction tasks. 
To establish a hierarchical joint representation,
Zhang et al. \cite{zhang2020multiscale} proposed a multi-scale adaptation fusion network (MAFN). The main contribution of MAFN is the adaptation fusion module (AFM) that incorporates features extracted from RGB and depth modalities and passes them to a neighbor attention module to enhance their local neighboring relational information. AFM is applied between the RGB and depth branches at multiple scales, as seen in Fig.~\ref{fig-afm}.

Li et al. \cite{cascade} introduced a cascaded hourglass network that consists of a branch (image encoder) used to extract features from images and three hourglass branches used to extract features from depth at different scales (1/4, 1/2, 1). The feature maps obtained from the image encoder at different scales are merged with the corresponding depth features by skip connection. The ground truth is down-sampled to different scales to make use of the multi-scale supervision. Such design enables a significant drop in model complexity and improves inference efficiency.

To better tackle the sparsity, many works seek to exploit additional constraints to guide the learning process. A common solution is to apply epipolar constraints between temporally adjacent frames \cite{wong2020unsupervised,wong2021adaptive,feng2022advancing,wong2021scaffnet,wong2021unsupervised,song2021self,choi2021selfdeco}, or stereo pairs \cite{ddp,shivakumar2019dfusenet}. 
Another constraint is adversarial loss which comes from adversarial training with the use of a generative adversarial network (GAN) \cite{Goodfellow2014GenerativeAN}. Although these constraints provide unsupervised guidance to the models for the depth completion task, they require additional inputs or other guidance networks during their training.

\subsubsection{Double encoder-decoder Networks}
\label{DEDN}

\textbf{Overall insight:} \textit{Extending dual-encoder networks to double encoder-decoder networks further boosts model performance.}

As discussed above, DEN-based methods usually consist of an RGB encoder, a depth encoder, and a decoder. The fusion is conducted between the two encoders.  
A double encoder-decoder network (DEDN) is an improvement of the dual-encoder network. A vanilla DEDN contains two encoder-decoder networks. In like manner, one takes an image input, and the other takes sparse depth input. The image network is also called the guided network. For methods built on DEDN, the fusion is usually conducted between the decoder of the image branch and the encoder of the depth branch at multi-scales.

\begin{figure}[h]
    \centering
    \includegraphics[width = 0.9\linewidth]{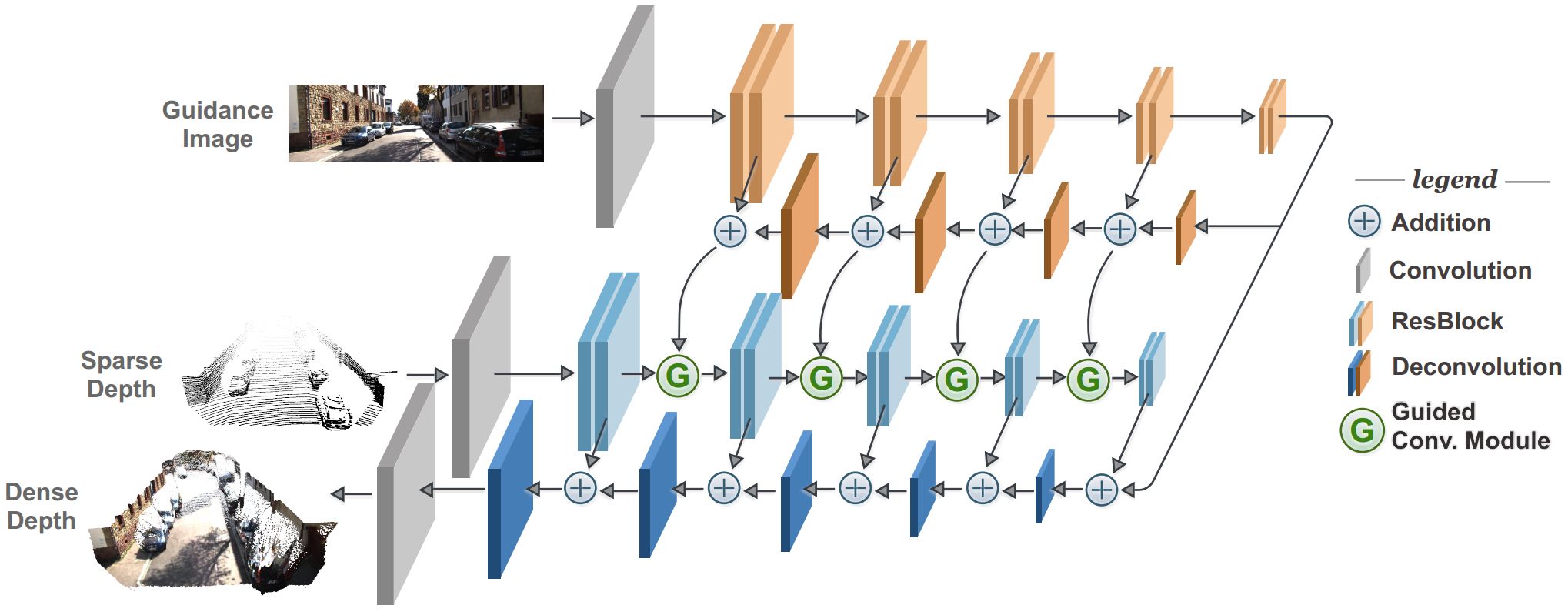}
    \caption{The architecture of the GuideNet. The framework is a double encoder-decoder network in which the guided convolution learns fusion kernels from RGB features and applies them to depth features.
    From \cite{learning-guided}.}
    \label{fig-guided-cnn}
\end{figure}

As a representative method depicted in Fig.~\ref{fig-guided-cnn},
GuideNet \cite{learning-guided} aims to learn a more effective fusion of RGB and depth features.
Inspired by guided image filtering \cite{guided-image-filtering} and bilateral filtering \cite{bilateral-filtering}, GuideNet introduced the guided convolution which automatically generates spatially-variant kernels from the image features and applies them to assign weights to the depth features.  The guided convolution is applied to multi-scale image features.
To reduce the computational complexity, motivated by MobileNet-V2 \cite{mobilenetv2}, the guided convolution is factorized into a channel-wise and a cross-channel convolution.

 Inspired by \cite{learning-guided} and \cite{SICNN},
 Schuster et al. \cite{schuster2021ssgp} proposed sparse spatial guided propagation (SSGP) which combines image guided spatial propagation and sparsity convolution. SSGP is applicable to not only depth completion but also other interpolation problems such as optical flow and scene flow.  Since SSGP aims to generalize to several visual tasks, its model design lacks focus on depth completion, resulting in inferior performance compared to the GuideNet.
More recently, Yan et al. \cite{yan2021rignet} proposed RigNet with a novel repetitive design to handle blurry object boundaries and better recover scene structures. In RigNet, the branch used for extracting image features is implemented using a repetitive hourglass network (RHN), i.e., multiple encoder-decoder networks, to produce perceptually clear image features. The branch of RigNet used for extracting depth features is also a hourglass network stacked with a repetitive guidance module (RG). RG plays a similar role as the guided convolution \cite{learning-guided} and is built on dynamic convolution \cite{chen2020dynamic}.
Since RG implements dynamic convolution repetitively, the convolution factorization proposed in \cite{learning-guided}  becomes less efficient. 
Thus, they designed an efficient guidance algorithm in which the kernel size in the channel-wise convolution drops from 3$\times$3 to 1$\times$1 by using global average pooling.  
RigNet achieves an extraordinary performance and currently ranks second on the KITTI depth completion dataset \cite{SICNN}.

\subsubsection{Global and Local Depth Prediction}
\label{GLDPN}

\textbf{Overall insight:} \textit{GLDPN takes advantage of both early fusion and late fusion by using a global depth prediction network and a local depth estimation network.}

In several prior works, 
 RGB and LiDAR data are referred to as global information, and the LiDAR data is referred to as local information. The global and local depth prediction (GLDP) methods employ a global network to infer depth from global information (global information is equivalent to early fusion of RGB images and sparse depths) and a local network to estimate depth from local information. The final dense depth map is obtained by merging the outputs of the global and local networks. 

To exploit both the global and local features, a global depth and local depth map, as well as related confidence maps, were predicted in \cite{sparse-and-noisy}. The confidence map predicted at each branch was used as a cross-guidance to refine the depth map predicted by the other branch. 
A similar method was also introduced in \cite{lee2020deep} where Lee et al. made two improvements. First, in order to extend the receptive field, they designed a residual atrous spatial pyramid (RASP) block to replace the traditional residual block. Second, unlike \cite{sparse-and-noisy} where the confidence map was directly used to refine a depth map via element-wise multiplication, they introduced a new guidance module that applies both channel-wise and pixel-wise attention operations.
The same framework was likewise used to address depth completion from the extremely sparse depths in order to explore depth completion from single-line depth maps in \cite{lu2021sgtbn}.

\subsubsection{Discussion} 
Late fusion models are more complex in the design of network architecture and multi-modality data fusion than early fusion models. Besides, they usually outperform in accuracy. 

Since RGB and depth are considered separate inputs for late fusion models, two separate convolutional modules are required for feature extraction. Therefore, it is quite instinctive to use DENs as proposed in existing studies. Then, naturally, the performance can be further improved by extending DENs to DEDNs. 
Second, we can also see a simple to complex evolution while exploring the process of developing feature fusion strategies. Multi-modal feature fusion methods have advanced from
 direct concatenation \cite{max-S-and-D} or summation \cite{shivakumar2019dfusenet,ryu2021scanline} in earlier works to applying semantic correlations \cite{zhong2019deep}, attention mechanisms \cite{zhang2020multiscale}, and spatially-variant kernels \cite{learning-guided} in more recent works;
from a single-spatial scale \cite{max-S-and-D,shivakumar2019dfusenet,ryu2021scanline,zhong2019deep,fu2020depth} to a more common multiple spatial scales \cite{zhang2020multiscale,cascade,learning-guided,yan2021rignet}. 

For the three types of late fusion models, both GLDP and DEDN are considered an improvement of early methods from the aspect of model design.
GLDP is a combination of early fusion methods with an additional network that predicts a dense depth map from a sparse depth input, while DEDN is an improvement of DEN and applies separate encoder-decoder networks for both an RGB and a sparse depth map input. Admittedly, DEDN will perform better than DEN if they are built on the same backbone network. On the other hand, although DEDN beats GLDP on benchmark datasets, they tend to use more complicated networks with much more parameters, e.g., GuideNet has 62.6M parameters while methods of GLDP including \cite{sparse-and-noisy} and \cite{lee2020deep} only yield 2.5M and 5.4M parameters, respectively.
Overall, \cite{sparse-and-noisy,cascade} 
achieve the best accuracy-efficiency trade-off.

\subsection{Explicit 3D Representation Models}
\label{E3DR}

Most previous studies of RGB guided depth completion learn 3D geometric relationships in an implicit yet ineffective manner. Typically, the difficulty comes from the incapability of normal 2D convolution to capture the 3D geometric clues from the sparse input where the observed depth values are irregularly distributed.
Hence, another type of previous approaches promotes explicit 3D representations (E3DR).
Previous methods of this type can be classified into the methods employing 1) 3D-aware convolution, 2) intermediate surface normal representation, and 3) methods of learning geometric representations from point clouds.

\subsubsection{3D-aware Convolution}
\label{3d-awae}

\textbf{Overall insight:} \textit{Since a depth point is correlated to its spatial neighbors, and there are many  missing points irregularly distributed in the sparse input, instead of the standard convolutions, applying 3D-aware convolutions to the nearest neighbors of a depth point helps eliminate perturbations of missing values.}

In 2D-3D FuseNet \cite{2d-3d}, features extracted from an RGB branch and a depth branch are fused by several 2D-3D fusion blocks that jointly learn 2D and 3D representations. The 2D-3D fusion block uses a multi-scale branch to extract appearance features in 2D grid space with normal convolution operations, and a branch to learn 3D geometric representations by applying two continuous convolutions \cite{wang2018deep} on K-nearest neighbors of a center point in 3D space.
 The idea of learning from spatially close K-nearest neighbors is then commonly employed in subsequent studies.
 
\begin{figure}[h!]
    \centering
    \includegraphics[width = 0.9\linewidth]{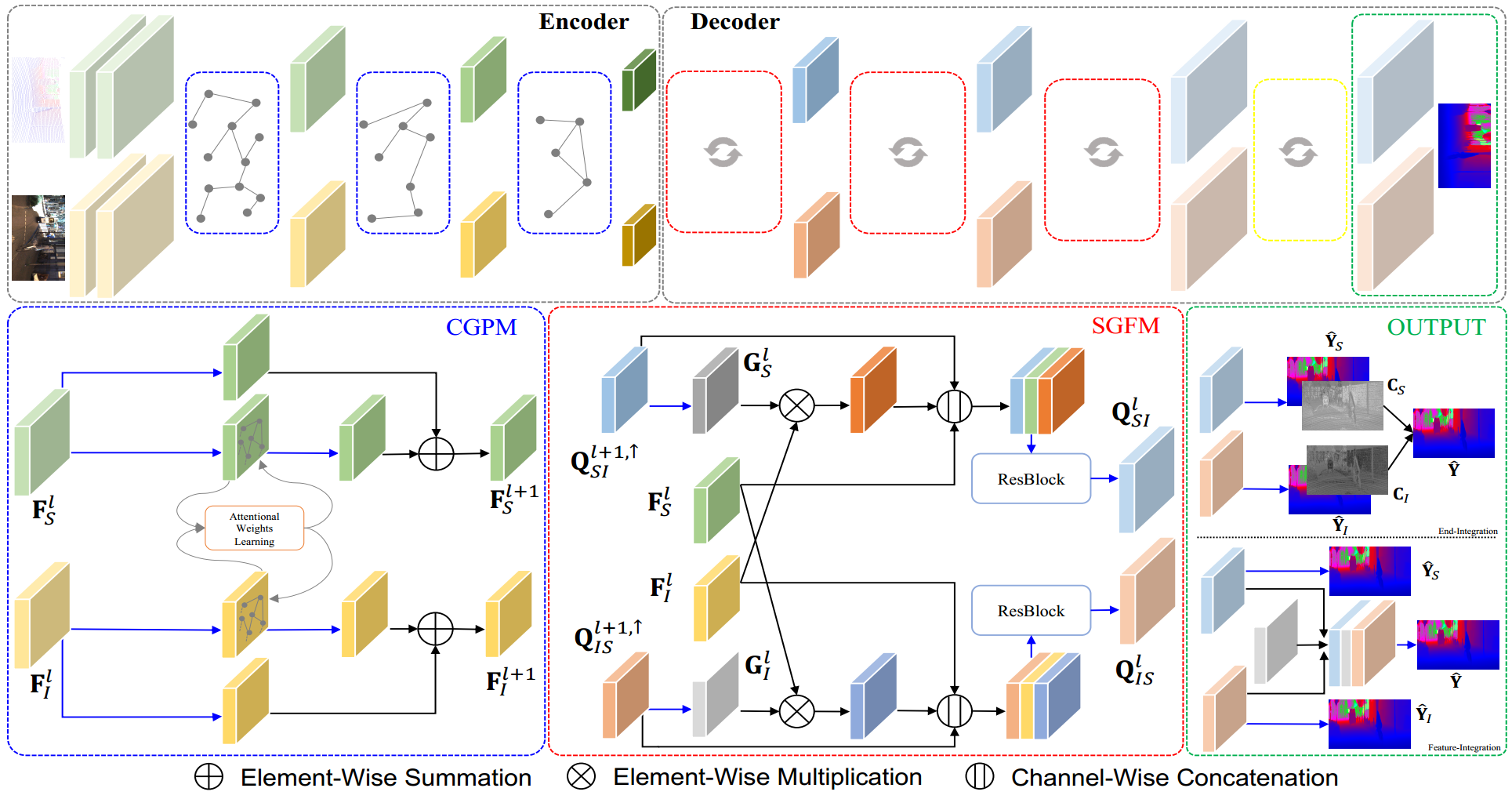}
    \caption{The diagram of the ACMNet where the encoder uses several co-attention guided graph propagation modules (CGPMs) for multi-modality feature extraction and the decoder uses several symmetric gated fusion modules (SGFMs) for multi-modality feature fusion.
    From \cite{zhao2021adaptive}.}
    \label{fig-acmnet}
\end{figure}

 For instance, in the ACMNet \cite{zhao2021adaptive}, the nearest neighbors are identified similarly by comparing the spatial differences. 
Unlike \cite{2d-3d}, the non-grid convolution is implemented by graph propagation. As seen in Fig.~\ref{fig-acmnet},
ACMNet has a DEDN structure where the encoder is composed of co-attention guided graph propagation modules (CGPMs), and the decoder is a stack of symmetric gated fusion modules (SGFMs). 
CGPM adaptively applies attention based graph propagation in both the image and depth encoders for multi-modality feature extraction, and SGFMs apply symmetric cross guidance between two decoders for multi-modality feature fusion.

Xiong et al. \cite{Xiong2020SparsetoDenseDC}  considered a graph model for depth completion and introduced a graph neural network (GNN) based depth completion algorithm. Note that the 3D graph of nearest neighbors is only constructed for a valid point in \cite{2d-3d,zhao2021adaptive}, while it is constructed for each point from a dense depth pre-enhanced from a baseline model with a DEDN architecture in \cite{Xiong2020SparsetoDenseDC}. 
In addition, it is worth mentioning that the method also studied and compared different sampling strategies for synthesizing sparse depth maps on the benchmark NYU-v2 dataset. The results show that quasi-random sampling \cite{niederreiter1992random} significantly outperforms random sampling\footnote{Previous methods of depth completion commonly validate their effectiveness on the NYU-v2 by synthesizing sparse depth maps via randomly sampling a few depth points.}. These findings can help perform experiments of different sampling strategies on the indoor datasets for the depth completion task.

\subsubsection{Intermediate Surface Normal Representation}
\label{surface-normal}

\textbf{Overall insight:} \textit{Surface normal is commonly used as an intermediate representation, and effective for indoor depth enhancement. Intuitively, is it also applicable for outdoor depth completion?}

\begin{figure}[h]
    \centering
    \includegraphics[width = 0.9\linewidth]{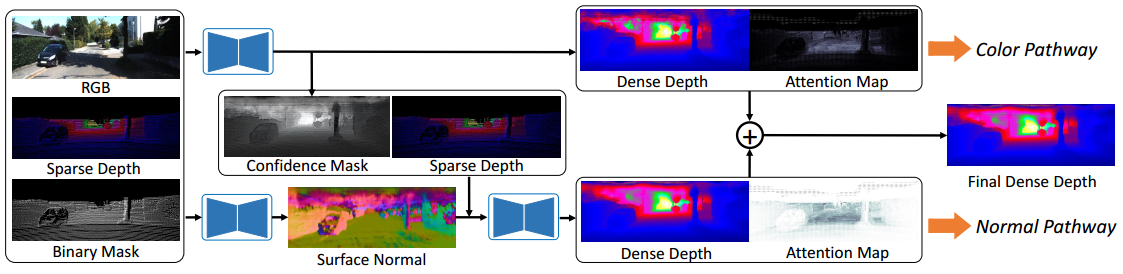}
    \caption{The pipeline of the DeepLIDAR where surface normal is used as an intermediate representation of a depth map. From \cite{DeepLiDAR}.}
    \label{fig-surface-normal}
\end{figure}

A few works utilized surface normal as an intermediate 3D representation of depth map and introduced methods employing surface normal guided completion.
As studied in \cite{DDC_of_single,huang2019indoor}, surface normal is a reasonably intermediate representation and can promote indoor depth enhancement. However, as pointed out by Qiu et al. \cite{DeepLiDAR} that reconstructing depth from normal in outdoor scenes is more sensitive to noise and occlusion; how to utilize surface normal in this case is still an open question. To address this issue,  they proposed DeepLIDAR, a two-branch network consisting of a color pathway and a surface normal pathway depicted in Fig.~\ref{fig-surface-normal}. 
Both branches produce a dense depth map. The final depth map is obtained via the attention-based weighing of the outputs of the two pathways.
 In the surface normal branch, surface normal is utilized as the intermediate representation of the produced depth map. 

The use of surface normal is straightforward for the method proposed in \cite{DeepLiDAR}. 
 As argued in \cite{xu2019depth}, the relation between depth and surface normal can be established via the tangent plane equation in the camera coordinate system. By this intuition, Xu et al. \cite{xu2019depth} proposed the plane-origin distance that forces the consistency between depth and surface normal to regularize depth completion.  Different from \cite{DeepLiDAR}, the method also estimates a confidence map modeling as Laplace distribution to mitigate the effect of noise and applies a refinement network. Benefiting from the depth and normal consistency, they achieved comparable performance against \cite{DeepLiDAR} while using only about 20$\%$ of parameters.

\subsubsection{Learning from Point Clouds}
\label{point-clouds}

\textbf{Overall insight:} \textit{We can explicitly extract 3D cues through directly learning from point cloud, since it is a reliably strong prior of 3D structures.}

Recently, a few studies directly learned geometric representations from point clouds. For example, Du et al. \cite{Du2022DepthCU} proposed to firstly learn a geometric-aware embedding from point clouds with edge convolution \cite{Wang2019DynamicGC}. Then, a DEN was utilized to perform depth completion from RGB images and geometric embeddings. 
 Jeon et al. \cite{jeon2021abcd} also used a point cloud as input. By incorporating the attention mechanism into bilateral convolution \cite{su2018splatnet}, they designed an attention bilateral convolutional layer (ABCL) based encoder for feature extraction from 3D point clouds. Their framework also implements a DEN where a point cloud encoder is used to extract 3D features, and an image encoder is used to extract 2D features from an RGB image and a sparse depth input.
 
As shown in \cite{jeon2021abcd,Du2022DepthCU}, integrating point cloud into depth completion significantly boosts the model generalization accuracy in different environments. Compared to \cite{jeon2021abcd}, the method of \cite{Du2022DepthCU} achieves competing results with yet a simple and more lightweight framework.

\subsubsection{Discussion}
Overall, methods employing E3DR outperform most approaches without explicit 3D representations, such as EDN, C2RP, and DEN. 
For these three explicit 3D models, methods employing 3D-aware convolution (3DAC) such as 2D-3D FuseNet \cite{2d-3d} and ACMNet \cite{zhao2021adaptive} outperform those using intermediate surface normal representation (ISNR) and learning from point cloud (LfPC) in both accuracy and efficiency.
It is not surprising because 3D-aware convolution only acts on spatially close valid depth points and is thus less influenced by missing values and helps reduce redundancy. 

 In general, 2D-3D FuseNet outperforms ACMNet in model complexity and underperforms in inaccuracy. The network used in \cite{Xiong2020SparsetoDenseDC} consists of a DEDN and a GNN module. Therefore, we believe that its model complexity is higher than those proposed in \cite{2d-3d,zhao2021adaptive}. 
However, since the implementation details, including the number of parameters, results on the official KITTI test set, etc., are unclear, we are not able to compare them in detail.
Nevertheless, as shown in \cite{Xiong2020SparsetoDenseDC}, we find that this method demonstrates a similar performance to \cite{xu2019depth} while the latter method is inferior to \cite{2d-3d,zhao2021adaptive}.

On the other hand, ISNR demonstrates performance comparable with 3DAC and LfPC while yielding the highest model complexity. It is because noise observed in outdoor scenarios will cause large distance errors even if they demonstrate small surface normal errors. Although the noise effect has been reduced by applying an attention-based mask in \cite{DeepLiDAR}, or modeling confidence mask in \cite{xu2019depth}, employing intermediate normal representations essentially brings this difficulty into depth completion. Besides, generating trustable target normals for supervision is also more challenging in outdoor environments.
Current methods do not demonstrate many advantages considering their complexity and performance. In the future, further efforts need to be put into utilizing surface normal. For example, we can use a more noise-robust virtual normal loss as proposed in \cite{yin2019enforcing} to regularize scene structures. 

LfPC shows another approach for 3D-aware depth completion by learning directly from 3D point clouds. A point cloud is a strong prior that preserves 3D object information and is more robust to occlusions and illuminations.
 A clear advantage of LfPC is that it exhibits higher generalization accuracy compared to ISNR \cite{DeepLiDAR} and E3DR \cite{zhao2021adaptive} in diverse weather and light conditions, as shown in \cite{jeon2021abcd}. 
While various network designs have been proposed for RGB and sparse depth input, the current attempts to include point clouds are still simple yet insufficient, e.g., both \cite{jeon2021abcd} and \cite{Du2022DepthCU} adopt a common DEN framework.

\subsection{Residual Depth Models}
\label{RDM}

\textbf{Overall insight:} \textit{The inferred depth maps should be precise in overall structure and faithful in local detail. Hence, the one-stage prediction procedure can be decoupled into the estimation of a dense map and a residual map.}

Residual depth models (RDMs) predict a depth map and a residual map, and their linear combination obtains the final depth. Through the prediction of the residual map, the model can refine the blur depth prediction and yield finer results on object boundaries. 

\begin{figure}[h]
    \centering
    \includegraphics[width=0.9\linewidth]{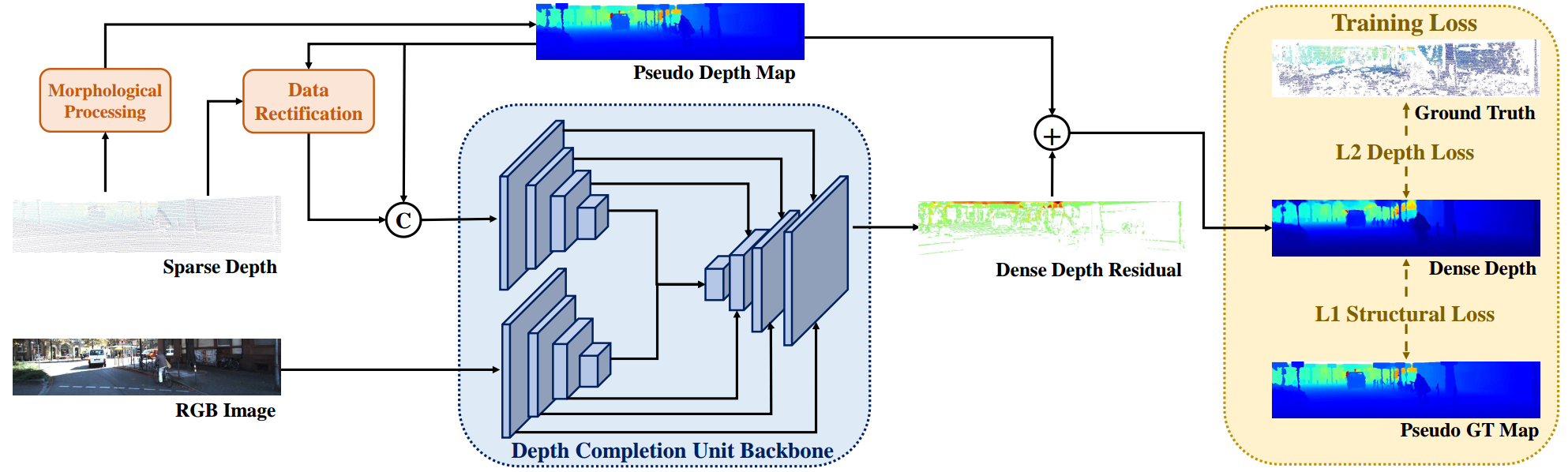}
    \caption{The pipeline of the DenseLiDAR where depth completion is decomposed as learning of a coarse depth map and a residual depth map. From \cite{gu2021denselidar}.}
    \label{fig_residul_learning}
\end{figure}

These methods usually apply a two-stage coarse-to-refinement likewise prediction procedure.
A simple application is shown in \cite{liao2017parse} where a sparse depth map is firstly completed to a dense map, and a residual map is then predicted. Finally, the element-wise summation of them generates the final depth map.
Gu et al. \cite{gu2021denselidar} proposed DenseLiDAR, a similar method as shown in Fig.~\ref{fig_residul_learning}. In DenseLiDAR, a pseudo depth map with morphological operations is firstly predicted. Then, the pseudo depth map, the RGB image, and the sparse depth input are sent to a CNN to predict a residual map. Finally, the pseudo depth map is rectified with the residual map to yield the final depth map.

For other approaches, the improvement is derived from
boosting the estimation of either the coarse depth map or the residual depth map.  
For instance, motivated by kernel regression, a differentiable kernel regression network was proposed to replace the hand-crafted interpolation for performing the coarse depth prediction from the sparse input in \cite{Nadaraya1964OnER,liu2021learning}. In addition, FCFR-Net \cite{fcfr} implemented an energy-based operation for multi-modal feature fusion to boost the residual map learning. 
  
Aiming at handling the uneven distribution and dealing with the outlier issue, Zhu et al. \cite{zhu2021robust} introduced a novel uncertainty based framework which consists of two networks: a multi-scale depth completion block and an uncertainty attention residual learning network. Like other residual based methods, the former network yields a coarse prediction, and the later network performs refinement. 
The uncertainty based framework prevents over-fitting from outliers by relaxing constraints of the highly uncertain regions in the first completion stage and guides the network to generate the residual map in the refinement stage.
 Zhang et al. \cite{zhang2021multi} combined the late fusion with residual learning and proposed a DEN-based multi-cue guidance network. Unlike other methods, the final depth is the combination of the sparse input and the estimated residual map.

\subsubsection{Discussion}
Residual depth models especially pay more attention to improving  the geometric fidelity of depth maps. Unlike C2RP, where refinement is directly performed from a pre-predicted coarse map, residual models apply residual learning to predict a residual map and use it as compensation for a dense map. The advantage of residual learning is typically reflected in two aspects. First, it can be regarded as a structural regulation and boosts perceptual quality. Second, since distance regions usually yield large depth errors and close regions have small depth errors, the residual learning tends to complement pixels with large errors and retains nearly zero values for near regions or perfectly pre-estimated pixels.

Overall, residual depth models achieve good completion performance. FCFR-net \cite{fcfr} and \cite{zhu2021robust} rank the fifth and ninth, respectively, on the KITTI benchmark dataset challenge. Note that the accuracy is not fully attributed to the residual learning, but also to the energy-based multi-modal feature fusion \cite{fcfr} and outlier-robust loss function \cite{zhu2021robust}. Since the number of parameters is unclear for these methods, we are not able to analyze their complexity.

\subsection{SPN-based Models}
\label{SPNM}

\textbf{Overall insight:} \textit{SPN explicitly forces  spatial correlations between a depth point and its neighbors via affinity-based refinement. }

An affinity matrix, also called a similarity matrix, expresses how close or similar data points are to each other. It is used to refine and gain a fine-grained prediction in vision tasks.
In spatial propagation networks (SPN) \cite{spn}, learning an affinity matrix is formulated as learning a group of transformation matrices. 
Following \cite{spn,nlspn}, the affinity refinement process of SPN is defined by
\begin{equation}
    x_{m,n}^t =  w_{m,n}^cx_{m,n}^{t-1} + \sum_{i,j \in \mathcal{N}_{m,n}}^{} w_{m,n}^{i,j}x_{i,j}^{t-1}
    \label{l1_affinity}
\end{equation}
where $(m,n)$ and $(i,j)$ denote the coordinates of reference and neighbor pixels, respectively, and $\mathcal{N}_{m,n}$ is a set of neighbor pixels of the reference pixel at $(m,n)$. $t$ denotes the iteration step of refinement.
$ w_{m,n}^c$ and $w_{m,n}^{i,j}$ are the affinity of the reference pixel and the affinity between the pixels at $(m,n)$ and $(i,j)$, respectively, where $w_{m,n}^c = 1 -  \sum_{i,j \in \mathcal{N}_{m,n}} w_{m,n}^{i,j}$.

\begin{figure}[h]
    \centering
    \includegraphics[width = 0.85\linewidth]{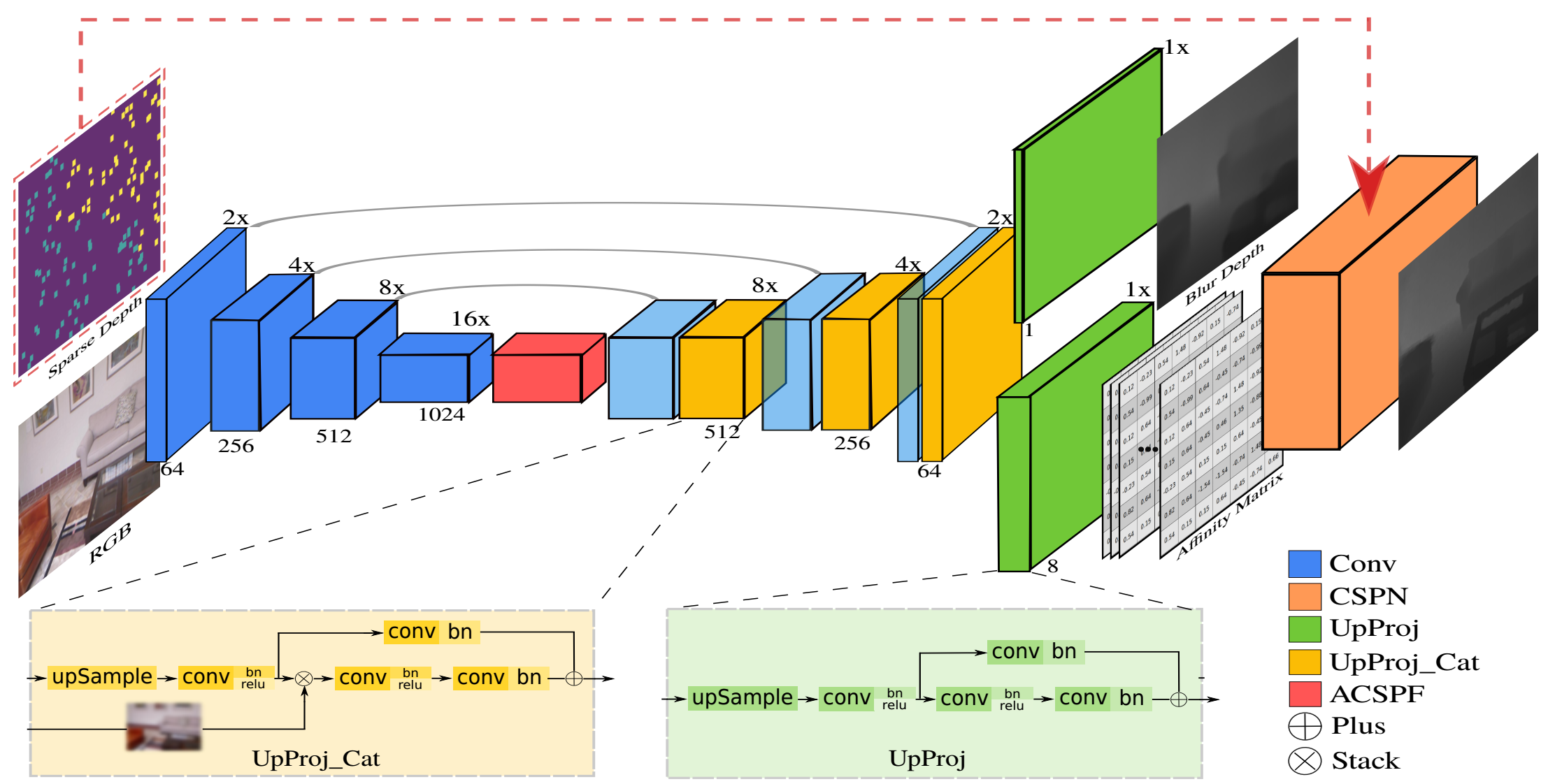}
    \caption{The framework of CSPN based depth completion. The CSPN module is plugged into the network to rectify a coarsely predicted depth map.
    From \cite{Cheng2020LearningDW}.}
    \label{fig_cspn}
\end{figure}

Since a depth point is correlated to its neighbors, the SPN is reasonably applicable to depth regression problems, and a family of previous studies developed their algorithms based on SPNs. Cheng et al. proposed the pioneering
convolutional spatial propagation network (CSPN) \cite{cspn,Cheng2020LearningDW} which is the first SPN-based model used for depth completion. 
Compared to the original SPN \cite{spn}, CSPN has two major improvements.
First, in SPN, a point is linked to three local neighbors from the nearest row or column, while in CSPN, a $3 \times3 $ local window is used to connect local neighbors. 
Second, CSPN efficiently propagates a local area in all directions via a convolution operation instead of propagating in different directions and integrating with max-pooling as SPN. 
The final value of a depth point is determined by its local neighbors via the diffusion process with the affinity matrix.
Specifically, the network proposed in \cite{S-D-single-image} is modified with skip connections and an additional output branch to generate the affinity matrix.
Given a coarse predicted depth map and the affinity matrix, a CSPN is plugged into the network \cite{S-D-single-image} for refinement, as shown in Fig.~\ref{fig_cspn}. 
The hyper-parameters including kernel size (size of local neighbors) and the number of iterations, need to be tuned by hyper-parameter search.

 To solve the difficulty of determining kernel sizes and iteration numbers,
  Cheng et al. further proposed CSPN++ \cite{cspn++} that enables an context aware CSPN (CA-CSPN) and an resource aware CSPN (RA-CSPN).
  For the implementation of CA-CSPN, various configurations of kernel sizes and numbers of iterations are first defined, and two extra hyper-parameters are introduced to weigh different kernel sizes and iterations adaptively.
  Thus, CA-CSPN consumes a large number of computational resources. 
 To tackle this issue, RA-CSPN selects the best kernel size and number of iterations for each pixel by minimizing the computational resource usage. To this end, a computational cost function is aggregated to the optimization target to balance the trade-off between accuracy and training time.

 While CSPN and CSPN++ mainly focus on the refinement from an existing encoder-decoder method \cite{S-D-single-image}, PENet \cite{penet} takes advantage of both SPN and late fusion models.
PENet uses the DEDN structure where one network predicts from RGB images and sparse depths, and the other network predicts from sparse depths and a pre-densified depth map. A CSPN++ is then applied to the fused depth map of these predictions.


The above methods use fixed local neighbors for spatial propagation during affinity learning. However, this will involve the unnecessary use of irrelevant local neighbors. To address this problem, Park et al. proposed a non-local SPN \cite{nlspn} where non-local neighbors with affinities and a depth confidence map are learned, and the propagation is implemented through the deformable convolutions \cite{zhu2019deformable} on the K non-local neighbors. Besides, they also designed the confidence-incorporated affinity normalization module to encourage more affinity combinations and reduce the negative effect of unreliable depth values.  


In \cite{dspn}, a deformable spatial propagation network (DSPN) is proposed to adaptively generate different receptive fields and affinity matrices for each pixel. 
Likewise, \cite{Lin2022DynamicSP} introduced attention based dynamic SPN (DySPN) 
that can learn an adaptive affinity matrix by decoupling neighboring pixels based on their distances. Such attention mechanism recursively generates different attention maps to refine the affinity matrix and bring us the new state-of-the-art method for depth completion. DySPN currently ranks first on the KITTI depth completion benchmark \cite{SICNN}.

\subsubsection{Discussion}

The first CSPN \cite{cspn} used a fixed $3\times3$ local neighbors and fixed kernel size. These two problems tend to involve irrelevant pixels and limit the representation capability of SPNs, respectively, and thus lead to the effects like over-smoothing.
The subsequent studies improved CSPN by handling these two problems, such as selecting non-local neighbors by Non-local SPN \cite{nlspn}, adaptively assigning pre-defined kernel sizes \cite{cspn++} by CSPN++ or designing attention based adaptive strategy by DySPN \cite{dspn}.

Thanks to these efforts, SPN-based models demonstrate a clear advantage over other types of methods in accuracy. Moreover, affinity-based refinement is not only more accurate as it explicitly applies geometry constraints in the depth space, but is also applicable to any of the existing models.

 However, as we also mentioned before, a method's advantage in accuracy is its disadvantage in complexity. SPN-based models inevitably need additional convolutional modules to implement SPN, and thus increase model complexity. In addition, affinity refinement is time-consuming due to multiple iterations of optimization. For instance, CSPN takes one second to complete a sparse map on the KITTI benchmark, which is $10-100$ times\footnote{Note that since utilized hardware platforms are different for these methods, we cannot obtain fully precise comparisons.} than most methods. For more recent methods, they are also inferior in inference efficiency even compared to other two-stage methods, e.g., DySPN and non-local SPN consume 0.16s and 0.2s, respectively, while FCFR-Net and DenseLiDAR spend 0.1s and 0.02s, respectively.

\section{Learning Objectives for Training Models}
\label{sec-loss}
Since depth completion and monocular depth estimation have the same target outputs, i.e., predicting dense depth maps, they share the same learning objectives, such as depth loss, surface normal loss, and photometric loss.
In this section, we describe the learning objectives used in previous studies. A brief overview is given in Table~\ref{table-loss} in which we will review the commonly used objectives in detail in the following sections. 

\subsection{Depth Consistency} \label{depth consistency}

Given a sparse input $Y'$, the predicted dense map $\hat{Y}$ where $\hat{Y}=N(Y';\mathcal{W})$, and the semi-dense ground truth depth map $Y$, 
many works \cite{max-S-and-D,ddp,long2021depth,tsuji2018non,shivakumar2019dfusenet} used the $l_1$ loss (mean absolute error) between the predicted depth map and the ground truth depth map on valid pixels by
\begin{equation}
    l_{1} = \frac{1}{n}\sum_{i=1}^{n}\| \hat{Y}_i - Y_i\|_1
    \label{l1_loss}
\end{equation}
where $\| \cdot \|_1$ denotes the $\ell_1$ norm, $\hat{Y}_i \in \hat{\mathcal{Y}}$ and $Y_i \in \mathcal{Y}$ denote the predicted depth and the ground truth depth at $i^{th}$ pixel, and  $n$ is the total number of valid depth points from $\mathcal{Y}$.
Also, most existing methods \cite{S-D-single-image,zhang2020multiscale,Du2022DepthCU} used the $l_2$ loss, also known as root mean squared error (RMSE) by
\begin{equation}
    l_{2} =  \frac{1}{n}\sum_{i=1}^{n}\|\hat{Y}_i -  Y_i\|_2,
    \label{l2_loss}
\end{equation}
where $\| \cdot \|_2$ denotes the $\ell_2$ norm.
Note that in many methods \cite{S-D-single-image,DeepLiDAR,S-D-single-image,S-d-selfsuper,lee2020deep}, the $l_2$ loss is referred to as MSE. Therefore, in this article, we do not technically distinguish between the RMSE and MSE when they are used as loss functions.

\begin{figure}[t]
    \centering
    \includegraphics[width = 0.75\linewidth]{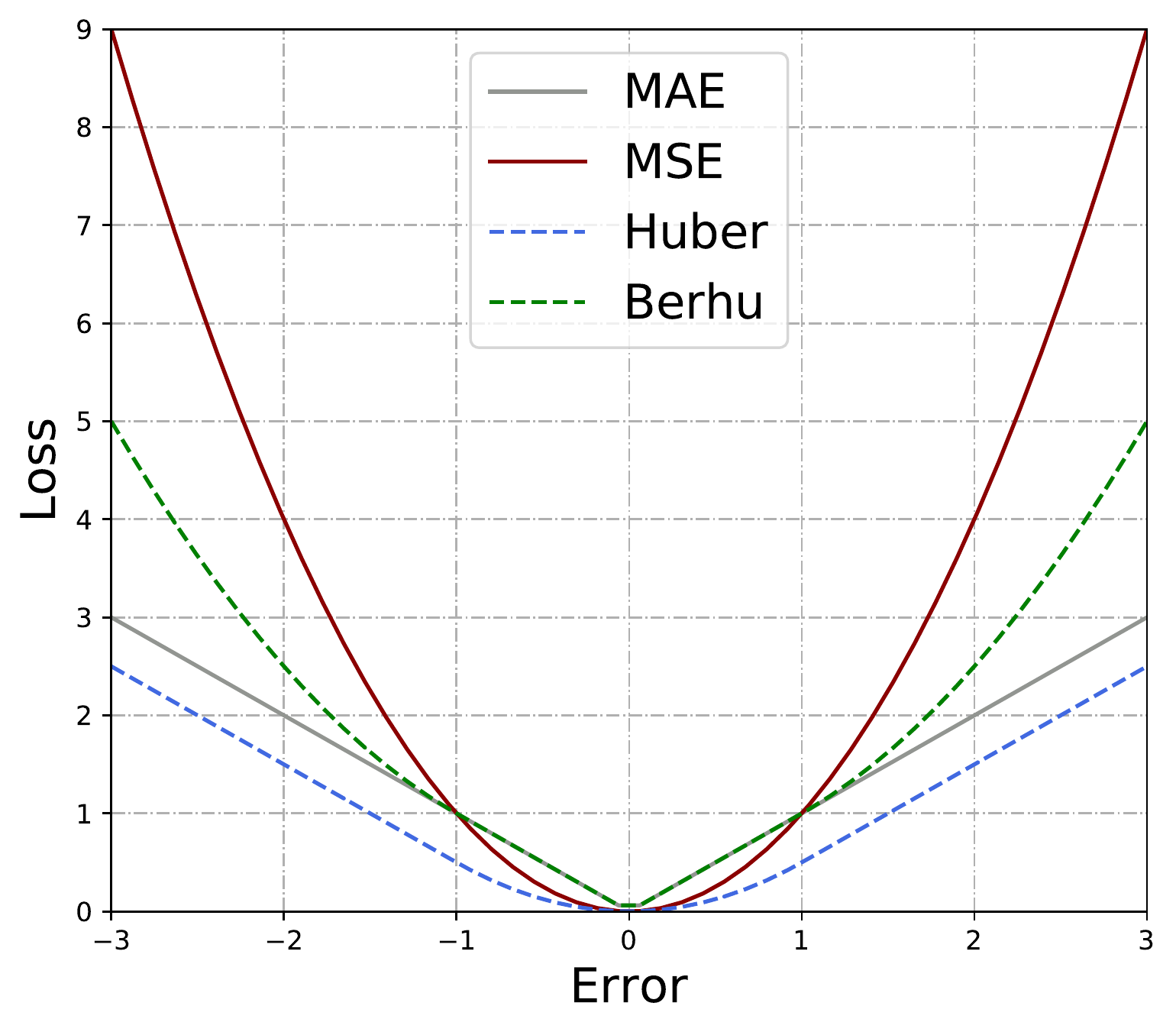}
    \vspace{-2mm}
    \caption{The comparison of MAE, MSE, Huber and Berhu norm.}
    \label{fig_vis_loss}
\end{figure}

The $l_{1}$ loss treats each valid pixel equally, while the $l_{2}$ loss is more sensitive to outliers and usually penalizes distant depth points more heavily. 
To take advantage of both losses, some methods attempt to combine them from different aspects. For example, several approaches \cite{hambarde2020s2dnet,Lin2022DynamicSP} linearly combined them as a loss function. 
Van Gansbeke et al. \cite{sparse-and-noisy} proposed focal-MSE where the mean absolute error was taken as a focal term for weighing the $l_2$ loss of depth.
Also, some works \cite{qu2020depth,eldesokey2019confidence} used the
Huber loss \cite{huber1992robust}  combining $l_{1}$ and $l_{2}$ to reduce the influence of large errors.  It is defined by
\begin{equation}
    l_{huber} =\left\{
    \begin{matrix}
         \frac{1}{n}\sum_{i=1}^{n} \ \frac{1}{2} (\hat{Y}_i -  Y_i)^{2}, &  |\hat{Y}_i - Y_i |\leq  \delta  \\ 
        \frac{1}{n}\sum_{i=1}^{n} \ \delta \left( |\hat{Y}_i - Y_i | -\frac{1}{2}\delta \right), & |\hat{Y}_i - Y_i | > \delta 
    \end{matrix}\right.
\label{huber-loss}
\end{equation}
where $|\cdot|$ denotes the absolute value operator and $\delta$ is usually set to 1. 
Besides, a few studies \cite{lopez2020project,eldesokey2019confidence} employ the Berhu loss \cite{owen2007robust_berhu} which is a reversion of Huber loss defined by
\begin{equation}
    l_{berhu}  =\left\{
    \begin{matrix}
     \frac{1}{n}\sum_{i=1}^{n} \  |\hat{Y}_i - Y_i|, & |\hat{Y}_i - Y_i| \leq \delta \\
 \frac{1}{n}\sum_{i=1}^{n} \  \frac{(\hat{Y}_i -  Y_i )^{2} + \delta^2}{2\delta },& |\hat{Y}_i - Y_i| > \delta 
    \end{matrix}\right.
\label{berhu-loss}
\end{equation}

Fig.~\ref{fig_vis_loss} visualizes the comparisons of MAE, MSE, Huber, and the Berhu loss functions for $\delta=1$. As shown, the Huber norm acts as $l_{2}$ when the error is less than $\delta$ and acts as $l_{1}$ otherwise. On the other hand, the Berhu norm acts inversely to the Huber norm, i.e., acts as $l_1$ when the error is less than $\delta$ and acts as $l_{2}$ otherwise.

Another attempt for handling the above issue of regression is to formulate depth prediction as a classification problem as an early work \cite{cao2017estimating} on monocular depth estimation. In this case, the depth range is discretized into a set of bins and a cross entropy loss is used. For depth completion, 
 \cite{depth_coefficient,liu2021learning} exploit this setting.

\begin{table*}[t]
\caption{A list of loss functions used for depth completion in previous works.}
\renewcommand\arraystretch{1.2}
\begin{center}
\scriptsize
\begin{tabular}
{|m{0.13\textwidth}<{\centering}|m{0.08\textwidth}<{\centering}|m{0.09\textwidth}<{\centering}|m{0.6\textwidth}<{\raggedright}|}
\hline
\textbf{Loss function} &\textbf{Type} & \textbf{Notation} &  \multicolumn{1}{c|}{\textbf{Explanation}} \\ \hline
\multirow{6}{*}{Depth Consistency} 
& \multirow{6}{*}{Supervised}  & $l_1$ & $l_1$ loss of depth on valid pixels, Eq.\eqref{l1_loss}  \\
& & $l_2$ & $l_2$ loss of depth on valid pixels, Eq.\eqref{l2_loss}.\\ 
& & $l_{huber}$ & Huber loss of depth on valid pixels, Eq.\eqref{huber-loss}.   \\ 
& & $l_{berhu}$ & Berhu loss of depth on valid pixels, Eq.\eqref{berhu-loss}.   \\ 
& & $l_{ce}$ & Cross entropy of depth on valid pixels by formulating depth regression as a classification task. \\
& & $l_{ud}$ & Uncertainty driven loss of depth on valid pixels, Eq.\eqref{uncertainty-loss2}.   \\
 \hline
\multirow{3}{*}{Structural loss} 
& \multirow{3}{*}{Supervised} & $l_{grad}$ & Gradient loss between the predicted depth map and the pseudo ground truth depth map.  \\
&  & $l_{normal}$ & Negative cosine difference of surface normal.  \\
&  & $l_{ssim}$ & SSIM loss  between the predicted depth map and the pseudo ground truth depth map. \\
\hline

 \multirow{3}{*}{ \makecell{Smoothness \\ regularization} }
 & \multirow{3}{*}{Unsupervised}
  & $l_{tv}$ \cite{chodosh2018deep} &Total variation of the predicted depth map.  \\
&  & $l_{smooth}$ & $l_1$ norm on second-order derivative of predicted depth map, Eq.\eqref{loss-smooth-1} or edge-aware smoothness loss,  Eq.\eqref{loss-smooth-2}.\\ \hline

\multirow{3}{*}{Geometric constraint} 
 & \multirow{3}{*}{Unsupervised}  &$l_{photo}$ & Photometric loss derived from temporally adjacent images or stereo images,  Eq.\eqref{photometric-loss}.\\ 
&  & $l_{stereo}$ & $l_2$ loss of depth between the predicted depth map and the pseudo ground truth depth map generated from stereo images.  \\ \hline

Adversarial loss
&Unsupervised & $l_{adv}$ & Adversarial loss  between the predicted depth map and the pseudo ground truth depth map,  Eq.\eqref{adv-loss}. \\ \hline

\multirow{6}{*}{Others} 
 & \multirow{6}{*}{Supervised}
  & $l_{tp}$ \cite{wong2021scaffnet}  &  $l_1$ loss between the prior (initial) depth map and the final estimated depth map.  \\
&  & $l_{cpn}$ \cite{ddp}& $l_2$ loss between an estimated depth map and its reconstruction from the conditional prior network.   \\
&  & $l_{cosine}$  \cite{long2021depth} & Cosine similarity between the predicted depth map and the pseudo ground truth depth map.  \\ 
&  & $l_{conf}$  \cite{xu2019depth} & Loss for learning the confidence map.  \\ 
&  & $l^{img}_p$  \cite{from_depth_what} & $l_p$ Loss for image reconstruction. \\ 
&  & $l_{ur}$ \cite{zhu2021robust} & Uncertainty aware loss for learning the residual depth map.   \\
\hline
\end{tabular}
\end{center}
\label{table-loss}
\end{table*}

Besides the above discussed loss functions, to tackle the outliers and inherent noises of the sparse input,
uncertainty aware learning objectives are also exploited.
 Uncertainty estimation \cite{kendall2017uncertainties} has been originally proposed to improve the robustness and accuracy of deep models.
Inspired by \cite{kendall2017uncertainties}, a couple of methods \cite{eldesokey2020uncertainty,zhu2021robust} introduce the uncertainty driven depth loss function where the completion is posed as maximizing the posterior probability.
Assuming the likelihood term $p(\hat{Y}_i|\sigma_i,Y_i)$ is modeled by a Gaussian distribution,
following \cite{eldesokey2020uncertainty,zhu2021robust}, 
then
\begin{equation}
    p(\hat{Y}_i|\sigma_i,Y_i) \approx  \frac{1}{\sqrt{2\pi}\sigma_i} 
  \exp\left( -\frac{(\hat{Y}_i-Y_i)^{\!2}}{2\sigma_i^{\!2}}\,\right)
\end{equation}
$\hat{Y}_i$ and $\sigma_i$ can be obtained via maximum likelihood estimation by
\begin{equation}
 \begin{aligned}
   \hat{Y}_i,\sigma_i & = \operatorname*{argmax}_{\hat{Y}_i,\sigma_i} \log p(\hat{Y}_i|\sigma_i,Y_i) \\ 
 & = \operatorname*{argmax}_{\hat{Y}_i,\sigma_i}  -\frac{1}{2}\log(2\pi) - \log(\sigma_i) -\frac{(\hat{Y}_i-Y_i)^{\!2}}{2\sigma_i^{\!2}} \\
 & =  \operatorname*{argmax}_{\hat{Y}_i,s_i}  -\frac{1}{2}\log(2\pi) - \frac{1}{2}\log(s_i) -\frac{(\hat{Y}_i-Y_i)^{\!2}}{2s_i} \\
 \end{aligned}
 \label{uncertainty-loss0}
\end{equation}
where $s_i \triangleq \sigma_i^{\!2}$ denotes the uncertainty of prediction at the $i^{th}$ pixel. Given equation ~\eqref{uncertainty-loss0}, the uncertainty driven depth loss for depth completion is defined by
\begin{equation}
     l_{ud}  =
    \frac{1}{n}\sum_{i=1}^{n}\left(\frac{(\hat{Y}_i-Y_i)^{\!2}}{s_i} + \log(s_i)\right)
	\label{uncertainty-loss}
\end{equation}
 In practice, an exponential function is usually applied to avoid division by zero during the training and the following uncertainty aware learning objective is used instead:
\begin{equation}
    l_{ud} = \frac{1}{n}\sum_{i=1}^{n} (\exp^{-s_i}(\hat{Y}_i-Y_i)^{2} + s_i).
	\label{uncertainty-loss2}
\end{equation}
In both works \cite{eldesokey2020uncertainty,zhu2021robust}, the uncertainty map $s$ is estimated with an additional branch within the depth completion framework.

\subsection{Structural Loss Functions}
\begin{figure}[t]
\centering
\includegraphics[width=80mm]{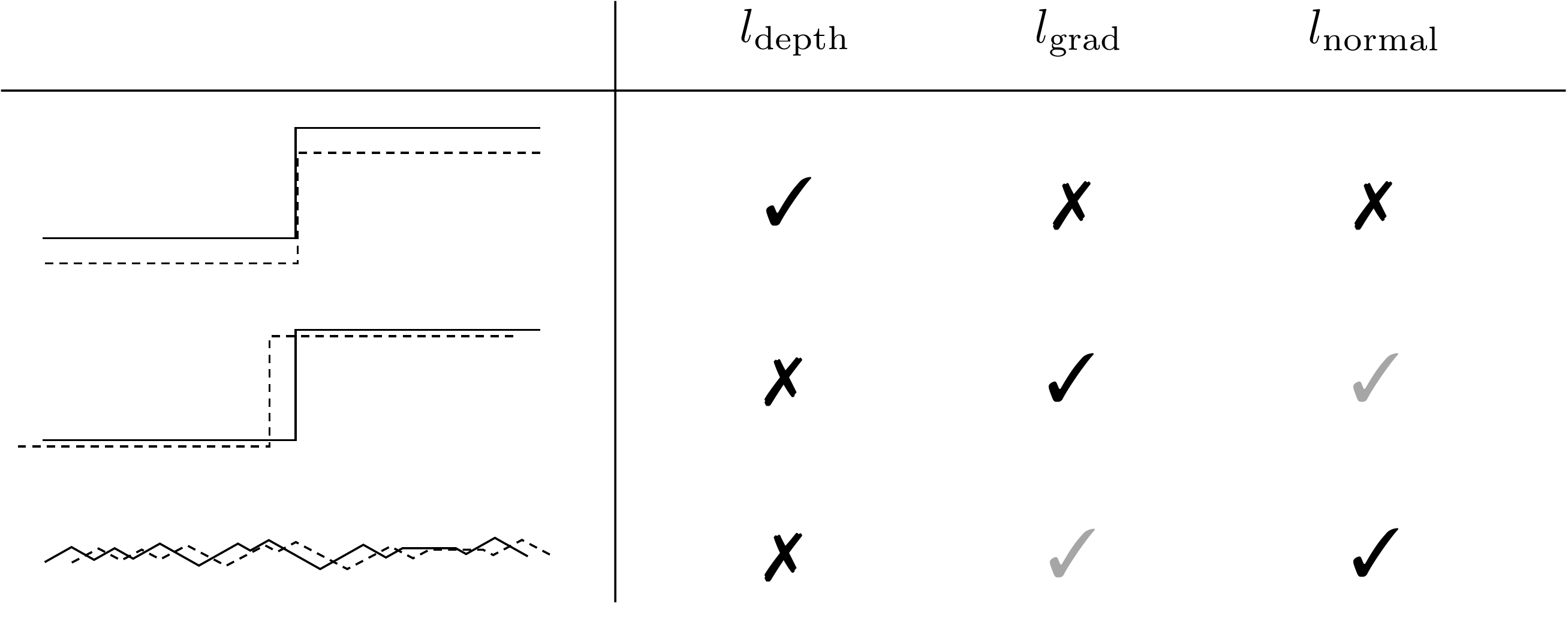}
\caption{Robustness of depth, gradient, and surface normal loss to depth differences. For simplicity, the solid and dotted lines denote two one-dimensional depth maps, respectively. It is observed that depth loss is insensitive to the shift and the occlusion of edges, while gradient and surface normal loss can handle these structural differences. From \cite{Hu2019RevisitingSI}.}
\label{fig_structural_loss}
\end{figure}

A common problem of previous works is that the predicted depth maps suffer from blur effects and distorted boundaries. To overcome this problem, researchers proposed to apply regularization to scene structures by introducing loss functions of depth gradient, surface normal, and perceptual quality.
Specifically, the gradient loss $l_{grad}$,  is implemented by minimizing the mean absolute error \cite{gu2021denselidar,liu2021learning}. For surface normal difference denoted by $l_{normal}$, the negative cosine difference is commonly utilized \cite{xu2019depth,DeepLiDAR}.
The effect of gradient and surface normal loss has been well studied in \cite{Hu2019RevisitingSI}. As shown in Fig.~\ref{fig_structural_loss}, the gradient loss contributes to penalizing errors emerging at the boundary of an object, while the surface normal loss can alleviate minor structural errors. 
Lastly, the structural similarity index measure (SSIM) loss \cite{ssim}, denoted by $l_{ssim}$, is penalized to ensure the perceptual quality  \cite{gu2021denselidar,multitask_gan}.
 Since dense ground truth depth maps are required, previous methods using the structural loss need to generate pseudo dense ground truth maps if they are not available from training data.

\subsection{Smoothness Regularization}
Smoothness regularization is utilized to suppress noises and ensure local smoothness for depth prediction.
There are typically two frequently used learning objectives for imposing depth smoothness. The first objective used in \cite{revisiting-scinn,S-d-selfsuper,zhang2019dfinenet,shivakumar2019dfusenet} is to minimize the $\ell_1$ norm on second-order derivative of predicted depth map by
\begin{equation}
    l_{smooth} = \frac{1}{n}\sum_{i=1}^{n}\left ( | \partial_{x}^{2} \ \hat{Y}_i | + | \partial_{y}^{2} \ \hat{Y}_i |\right )
    \label{loss-smooth-1}
\end{equation}
where $\partial_{x}$  and $\partial_{y}$ denotes the gradients along the horizontal and vertical direction of the dense depth map.
 The second is the edge-aware smoothness loss used in \cite{wong2020unsupervised,ryu2021scanline,wong2021scaffnet,wong2021unsupervised,song2021self,choi2021selfdeco} that allows depth discontinuity at boundaries by
\begin{equation}
    l_{smooth} = \frac{1}{n}\sum_{i=1}^{n}\left ( \left | \partial_{x} \hat{Y}_i \right | e^{-\left | \partial_{x}I_i\right |}+\left | \partial_{y} \hat{Y}_i \right | e^{-\left | \partial_{y}I_i\right |}\right )
    \label{loss-smooth-2}
\end{equation}

Besides, the total variation is also used in \cite{chodosh2018deep} for noise suppression.

\subsection{Multi-view Geometric Constraints}
One of the most challenging issues for depth completion is the lack of dense and high quality ground truth. 
To cope with this problem, researchers also attempt to seek solutions from the perspective of utilizing loss functions.
Among them, temporal photometric loss obtained from consecutive images provides an unsupervised supervision signal~\footnote{This signal is also called self-supervised signal in some studies \cite{ito2021seeing,S-d-selfsuper,song2021self}} to guide depth completion.

Ma et al. \cite{S-d-selfsuper} are the first that introduce photometric loss for depth completion. 
Based on the epipolar geometry, the predicted depth map of an image is warped to the nearby frame. Then, the differences at corresponding pixels are penalized.
Formally, given an image $I_t$ and its temporally adjacent image $I_{s}$, where $s \in \{t-1,t+1\}$, the warping of a pixel $p_i$ from $I_s$ to $I_t$ is computed by
\begin{equation}
    \hat{p}_i = KT^{s \rightarrow t}\hat{Y}(p_i) K^{-1}p_i
    \label{warping_function}
\end{equation}
where $K$ denotes the camera intrinsic matrix and $T^{s \rightarrow t}$ denotes the relative pose from $s$ to $t$. $\hat{Y}(p_i)$ is the predicted depth of the pixel $p_i$ of the image $I_s$.  $\hat{p}_i$ is the corresponding point of $p_i$ from $I_{t}$.

Then, 
the photometric loss between two images is defined by
\begin{equation}
    l_{photo} = \frac{1}{m}\sum_{i=1}^{m} \| I_s(p_i) - I_t(\hat{p}_i)\|_1
    \label{photometric-loss}
\end{equation}
where $m$ denotes the number of warped pixels.

Researchers attempt to improve the above photometric loss from different perspectives in subsequent studies.
The photometric loss is susceptible to moving objects. To alleviate this problem, Chen et al. \cite{chen2020spatiotemporal} integrated a MaskNet into the self-supervised framework. The MaskNet predicts the masks of moving objects such that the influence of moving objects can be reduced. Also, to ensure perceptual consistency, Wong et al. 
\cite{wong2020unsupervised,wong2021unsupervised} integrated the SSIM difference \cite{ssim} between warped and original images into the photometric loss. 

Different approaches to calculating the photometric loss have also been explored. In \cite{ito2021seeing}, optical flow is used to estimate the relative pose between two consecutive frames, and a pose estimation net is used for this purpose in \cite{zhang2019dfinenet}.
In \cite{song2021self}, relative poses are calculated in feature spaces at multiple scales.
Specifically, consecutive frames are sent to the FeatNet for multi-scale feature extraction. The relative pose is calculated with the Gauss-Newton algorithm \cite{NavarroPedreo1996NumericalMF} at each scale.

Wong et al. have put much effort into improving unsupervised depth completion.
As pointed out in \cite{wong2021adaptive}, the conventional use of the photometric loss treats each pixel equally. Unfortunately, this incurs significant meaningless errors at occluded regions. 
To address this issue, they proposed an adaptive weighting function \cite{wong2021adaptive} that acts as a flipped sigmoid function.
The weights for the photometric loss are approximately equal to 1 at each pixel at the beginning, and will get smaller if the residual at certain pixel increases during the training procedure.
In \cite{wong2020unsupervised}, the sparse depth is firstly densified by scaffolding operation \cite{brown1979voronoi,barber1996quickhull}, and then passed to an EDN for refinement. 
In ScaffFusion \cite{wong2021scaffnet}, the parameter-free scaffolding operation used in \cite{wong2020unsupervised} is replaced with a spatial pyramid pooling block as well as an encoder-decoder network. Then, ScaffFusion predicts a depth scale and a residual depth map for refinement.
To increase generalizability on different cameras, KBNet \cite{wong2021unsupervised} takes the calibration matrix as an additional input such that it can adjust to different cameras during inference.

A few previous works studied depth completion under the stereo setting except for the temporal photometric loss.
When a stereo pair is available, as seen in \cite{ddp}, the multi-view photometric consistency can be derived in a different fashion. 
Besides, in order to handle the lack of supervision, stereo images are used to generate ground truth depths for missing pixels in \cite{shivakumar2019dfusenet}. However, despite these advantages, the stereo setting inevitably lowers the generalizability of these methods \cite{ddp,shivakumar2019dfusenet} in practice.

\subsection{ Adversarial Loss}
Several approaches also adopt adversarial loss to promote depth completion \cite{tsuji2018non,to_complete_or_to_estimate,multitask_gan,khan2021sparse}. In these works, a generator is used to infer a depth map from the RGB and sparse depth map, and a discriminator is used to distinguish between the reconstructed depth map and ground truth by

\begin{equation}
\begin{gathered}
        l_{\rm adv} = \min\limits_{G}\max\limits_{D}\mathop{\mathbb{E}} [\log D( Y)] + 
        \mathop{\mathbb{E}} [\log(1-D(G(I, Y')))]
        \label{adv-loss}
\end{gathered}
\end{equation}
where $Y$ is dense ground truth which is usually obtained by other completion algorithms, $G$ and $D$ are the generator and discriminator, respectively.

\section{Datasets and Evaluation Metrics}
\label{sec-datasets}
In this section, we
introduce the benchmark datasets commonly used in previous works in detail. We also comprehensively survey the related datasets for reference. 

\subsection{Real-world Datasets}
\label{real-world datasets}
\subsection{Real-world Datasets}
\label{real-world datasets}
\textbf{KITTI depth completion dataset \cite{SICNN}: }  
The KITTI dataset is a widely used large-scale outdoor dataset that contains over 93,000 semi-dense depth maps with the corresponding raw sparse LiDAR scans and RGB images. The training, validation, and test set have 86,000, 7,000 and 1,000 samples, respectively.
The full resolution of images and depth maps can reach $1216\times352$, which is larger than most existing RGBD datasets. The raw LiDAR scans are captured by a Velodyne HDL-64E. To have a semi-dense ground truth depth map, Uhrig et al.~\cite{SICNN} purified the raw data with the semi-global matching (SGM) and densified the sparse depth map by accumulating 11 laser scans. 
 
\begin{figure}[h]
\centering  
\begin{tabular}
{p{0.131\textwidth}<{\centering}p{0.131\textwidth}<{\centering}p{0.131\textwidth}<{\centering}} 
\IncG [width=1.07in]{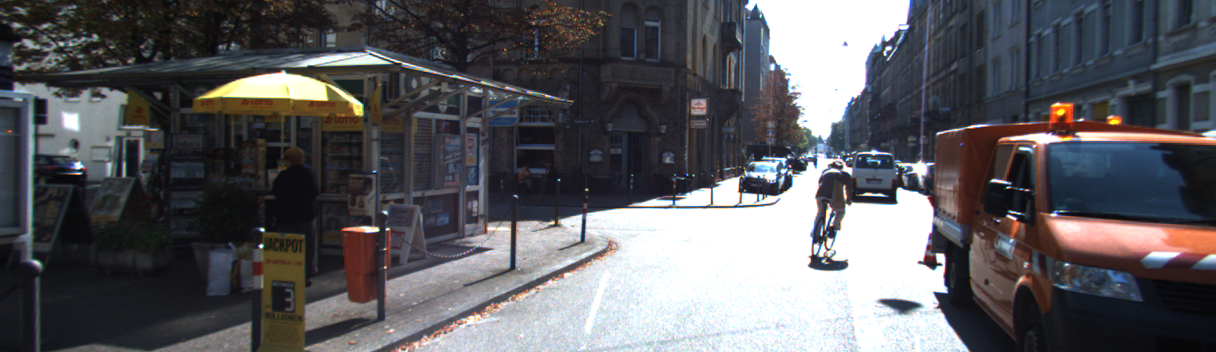} 
&\IncG [width=1.07in]{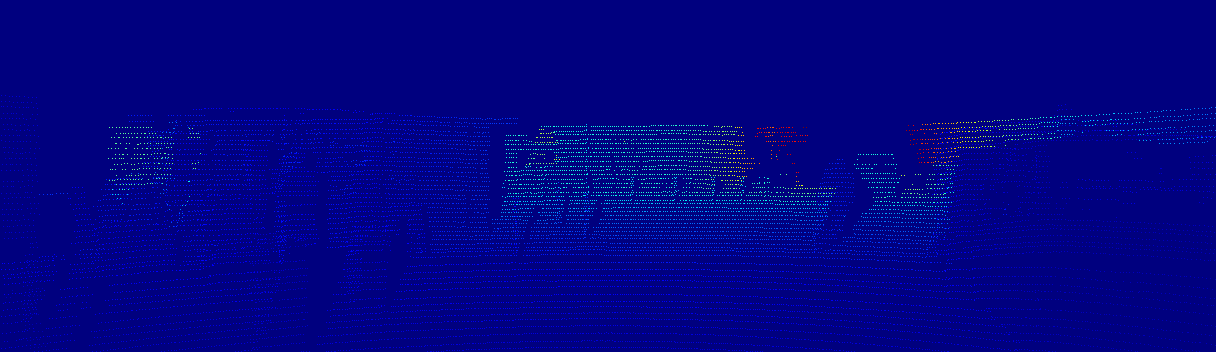}
&\IncG [width=1.07in]{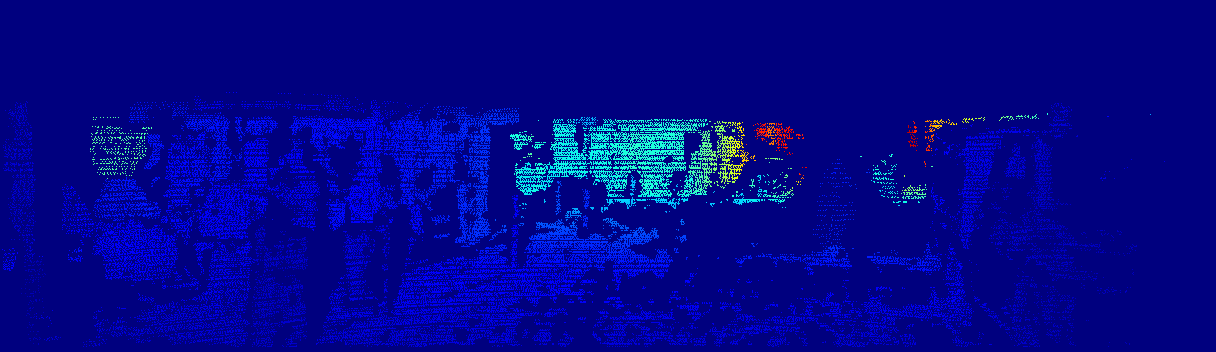}
\\
\IncG[ width=1.07in]{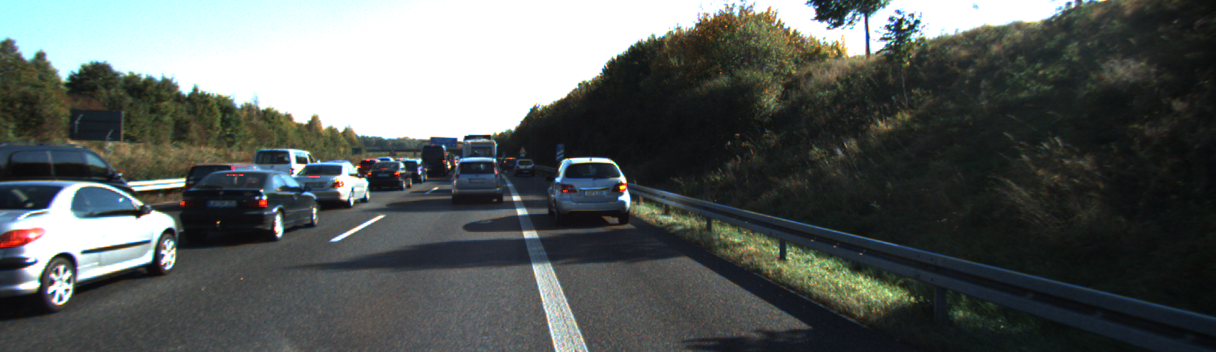} 
&\IncG[ width=1.07in]{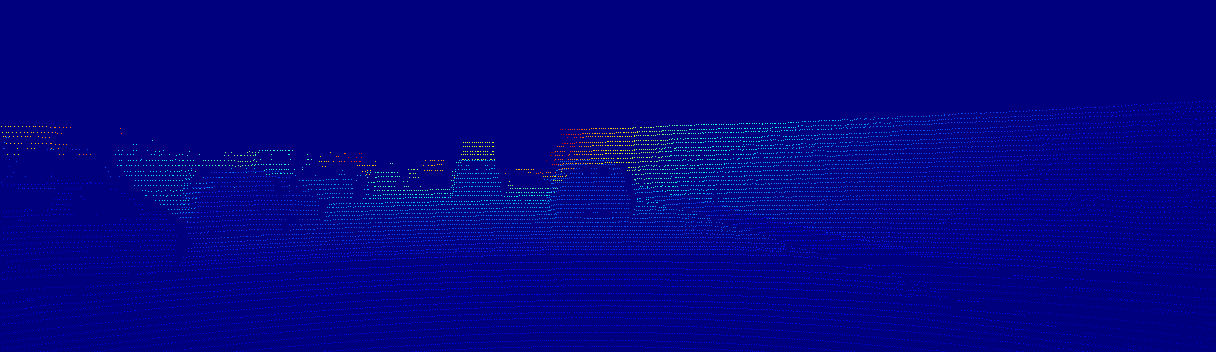}
&\IncG[ width=1.07in]{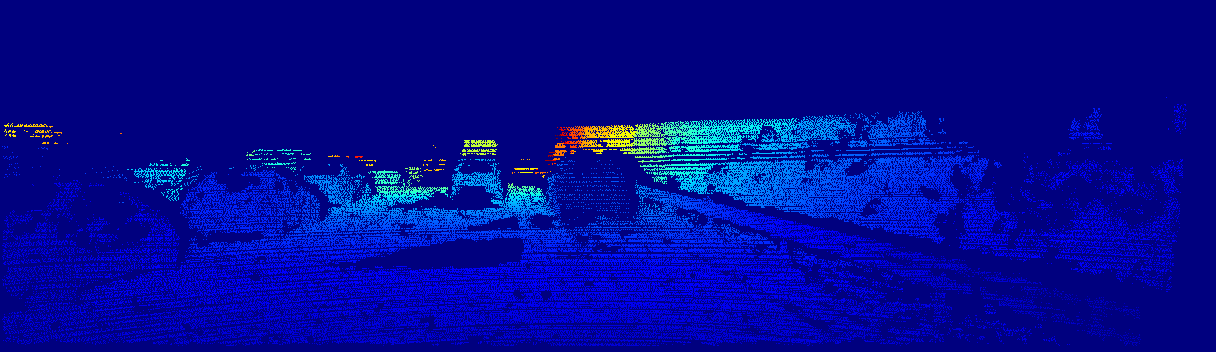}
 \\
 (a) &(b) &(c) \\
\end{tabular}
\caption{Sample images from the KITTI depth completion dataset \cite{SICNN}. (a) RGB images.  (b) Raw sparse depth maps. (c) Ground truth depth maps.}
\label{fig-kitti-samples}
\end{figure}

 It should be noted that the ground truths can be used differently in implementing previous methods. The density of the original sparse depth maps is only about $5\%$ (as observed in Fig.~\ref{fig-kitti-samples} (b)), and the semi-dense ground truths provided by the KITTI benchmark can reach about $30\%$ (as visualized in Fig.~\ref{fig-kitti-samples} (c)). 
Most previous works take the denser ground truths to implement their methods, whereas several unsupervised approaches \cite{ddp,wong2020unsupervised,wong2021scaffnet,wong2021adaptive,wong2021unsupervised} assume that only original sparse depth maps are available. In this case, the depth consistency is only applied to those $5\%$ valid pixels.

\begin{figure}[h]
\centering  
\begin{tabular}
{p{0.097\textwidth}<{\centering}p{0.097\textwidth}<{\centering}p{0.097\textwidth}<{\centering}p{0.097\textwidth}<{\centering}} 
\IncG [width=0.82in]{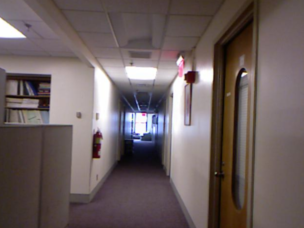} 
&\IncG [width=0.82in]{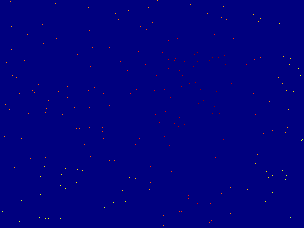}
&\IncG [width=0.82in]{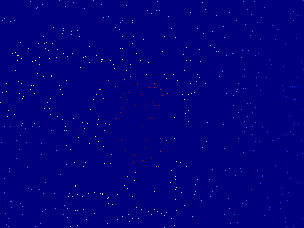}
&\IncG [width=0.82in]{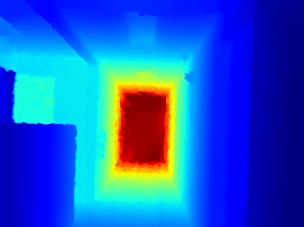}
\\
 (a) &(b) &(c) &(d)\\
\end{tabular}
\caption{A sample image from the NYU-v2 dataset \cite{NYUv2}. (a) An RGB image.  (b) A spare depth map (200 points). (c) A sparse depth map (500 points). (d) The corresponding ground truth depth map.}
\label{fig-nyu-samples}
\end{figure}

\textbf{NYU-v2 \cite{NYUv2}: } The NYU-v2 dataset consists of 464 indoor scenes with 408,000 RGBD images captured by Microsoft Kinect with an original resolution of $640 \times 480$. Although the original RGBD data is only applicable to depth enhancement methods, 
previous studies of depth completion implement their methods by randomly selecting 200 (Fig.~\ref{fig-nyu-samples} (b)) or 500 depth points (Fig.~\ref{fig-nyu-samples} (c)) as sparse inputs. The total valid pixels are less than $1\%$ in both cases.
Most methods evaluated on the NYU-v2 dataset are RGB guided.
 In the following sections, we show the essential characteristics of previous methods, including network structure, loss function, RMSE, etc., on the dataset.
 
 For methods evaluated on this dataset, unsupervised approaches can only apply depth consistency to those valid depth points in synthesized sparse depth inputs. In contrast, supervised methods such as \cite{S-D-single-image,cspn,cspn++} usually employ dense pixel-wise ground truth depth maps pre-densified by the official inpainting toolbox.  
 
\begin{table*}[t]
\caption{Summary of essential characteristics of existing unguided methods on the KITTI dataset. For denoting the loss function, we omit the coefficient of each loss term for simplicity. S and U denotes supervised learning and unsupervised learning of models, respectively.}
\renewcommand\arraystretch{1.1}
\begin{center}
\scriptsize
\begin{tabular}
{|m{0.115\textwidth}<{\raggedleft}m{0.05\textwidth}<{\centering}m{0.05\textwidth}<{\centering}|m{0.1\textwidth}<{\centering}m{0.16\textwidth}<{\centering}m{0.05\textwidth}<{\centering}|m{0.05\textwidth}<{\centering}m{0.05\textwidth}<{\centering}m{0.08\textwidth}<{\centering}m{0.03\textwidth}<{\centering}|}
\hline
\textbf{Method}  &\textbf{Publication}  &\textbf{Year} & \textbf{Type} & \textbf{Loss Function} &\textbf{Learning} & \textbf{RMSE (mm)}  &\textbf{Params (M)}  &\textbf{Platform} & \textbf{Code} \\ 
\hline
SI-CNN\cite{SICNN} &3DV & 2017  &SACNN   & $l_2$ &S &1601.33 &0.025 &TensorFlow &\checkmark  \\
DCCS\cite{chodosh2018deep} &ACCV & 2018  &SACNN &$l_2$ + $l_{tv}$ &S &1325.37 &0.0017 &TensorFlow &\checkmark \\
HMS-Net\cite{huang2019hms}  &TIP & 2019  &SACNN &$l_2$ &S  &937.48 &- &- &- \\
NConv-CNN\cite{Eldesokey2018PropagatingCT}  &BMVC & 2018 & NCNN  &$l_{berhu}$ &S &1268.22 &0.00048 &PyTorch &\checkmark \\
pNCNN\cite{eldesokey2020uncertainty} &CVPR    & 2020  & NCNN   &$l_{ud}$ &S  &960.05 &0.67 &PyTorch &\checkmark\\
DCAE \cite{Lu2022DepthCA} & WACVW & 2022 & TwAI &$l_1$ + $l_1^{img}$ &S\&U &1464.69 &2.29 &PyTorch &-\\
IR L1\cite{from_depth_what}  &CVPR    & 2020 & TwAI &$l_1$ + $l_2^{img}$ &S  &915.86 &11.63 &PyTorch &- \\  
IR L2\cite{from_depth_what}   &CVPR   & 2020 & TwAI &$l_2$ + $l_2^{img}$ &S &\textbf{901.43} &11.63 &PyTorch &- \\  
\hline
\end{tabular}
\end{center}
\label{table-method}
\end{table*}

\begin{table*}[t]
\caption{Summary of essential characteristics of selected existing RGB guided methods on the KITTI dataset. For denoting loss functions, we omit the coefficient of each loss term for simplicity. S and U denotes supervised learning and unsupervised learning of models, respectively. Accordingly, the top and bottom parts of the table show the supervised and unsupervised methods implemented for depth completion, respectively. }
\renewcommand\arraystretch{1.2}
\scriptsize
\begin{center}
\begin{tabular}
{|m{0.136\textwidth}<{\raggedleft}m{0.05\textwidth}<{\centering}m{0.05\textwidth}<{\centering}|m{0.1\textwidth}<{\centering}m{0.16\textwidth}<{\centering}m{0.05\textwidth}<{\centering}|m{0.05\textwidth}<{\centering}m{0.05\textwidth}<{\centering}m{0.08\textwidth}<{\centering}m{0.03\textwidth}<{\centering}|}
\hline
\textbf{Method}  &\textbf{Publication}  &\textbf{Year} & \textbf{Type} & \textbf{Loss Function} &\textbf{Learning} & \textbf{RMSE (mm)} &\textbf{Params (M)}  &\textbf{Platform} & \textbf{Code} \\ 
\hline
3coef\cite{depth_coefficient}&CVPR &2019 &EFM/EDN  &$l_{ce}$ &S &965.87 &- & TensorFlow & \checkmark \\
EncDec-Net[EF]\cite{eldesokey2019confidence} &TPAMI  & 2019 &EFM/EDN   &$l_1$ &S&965.45 &0.484 &Pytorch &\checkmark \\
Qu et al.\cite{qu2020depth} &WACV & 2020 &EFM/EDN   &$l_{huber}$ &S &998.80 &- &Pytorch &-\\
Morph-Net\cite{dimitrievski2018learningmorph} &ACIVS  & 2018 &EFM/C2RP &$l_2$ &S &1045.45 &- &Matlab &\checkmark\\
S2DNet\cite{hambarde2020s2dnet}  &TCI  & 2020 &EFM/C2RP   &$l_1 +  l_2$ &S&830.57 &- &PyTorch  &-\\
Long et al.\cite{long2021depth} & JVCIR  & 2021 &EFM/C2RP  &$l_1 + l_{cosine}$ &S &776.13 &- &- &- \\

Spade-RGBD\cite{max-S-and-D} &3DV &2018 &LFM/DEN   &iMAE &S &917.64 &5.3 &- &- \\
MS-Net[LF]\cite{eldesokey2019confidence}  &TPAMI & 2019 &LFM/DEN  &$l_1$ &S &859.22 &0.356 &PyTorch &\checkmark \\

MSG-CHN\cite{cascade}  &WACV & 2020 &LFM/DEN   &$l_2$ &S&762.19 &1.2 &PyTorch &\checkmark\\
MAFN \cite{zhang2020multiscale}   &IJCNN & 2020  &LFM/DEN &$l_2$ &S &803.50 &- &- &- \\  
Ryu et al.\cite{ryu2021scanline} &RAL &2021 &LFM/DEN & $l_2 + l_{smooth}$ &S &809.09 &1.9 &- &- \\
DVMN\cite{reichardt2021dvmn}      &ITSC & 2021 &LFM/DEN   & $l_2 + l_{smooth}$ &S&776.31 &- &- &-\\

GuideNet\cite{learning-guided} &TIP & 2020 &LFM/DEDN  &$l_2$ &S&736.24 & 62.62&PyTorch &\checkmark\\ 
SSGP\cite{schuster2021ssgp}    &WACV & 2021 &LFM/DEDN &$l_2$ &S&838.22 & 4.61 &- &-\\
RigNet\cite{yan2021rignet}    &Arxiv & 2021  & LFM/DEDN  &$l_2$ &S&712.66 & - &PyTorch &- \\

Van et al.\cite{sparse-and-noisy} &MVA & 2019 &LFM/GLDP &focal-MSE &S&772.87 &2.545 &PyTorch  &\checkmark\\
CrossGuidance\cite{lee2020deep}  &Access &2020 &LFM/GLDP &$l_2$ &S&807.42 &5.4 &PyTorch &-\\

2D-3D FuseNet\cite{2d-3d} &ICCV  &2019 &E3DR/3DAC  &$l_1 + l_2$ &S &752.88 &1.898 &- &- \\ 
ACMNet\cite{zhao2021adaptive}   &TIP & 2021  &E3DR/3DAC &$l_2$ &S &732.99 &4.9 &PyTorch &\checkmark \\

DeepLiDAR\cite{DeepLiDAR} &CVPR & 2019  &E3DR/ISNR &$l_2$ + $l_{normal}$ &S&758.38 &144 &PyTorch  &\checkmark\\
PwP\cite{xu2019depth}   & ICCV   & 2019  &E3DR/ISNR  &$l_2 + l_{normal} + l_{conf}$ &S&777.05 &28.99 &PyTorch  &- \\

ABCD\cite{jeon2021abcd}  &RAL & 2021  &E3DR/LfPC  &$l_2$ &S&764.61 &32.93 &PyTorch &- \\
Du et al. \cite{Du2022DepthCU} &Arxiv & 2022 &E3DR/LfPC &$l_2$ &S &773.90 &4.189 &PyTorch &\checkmark\\

FCFR-Net\cite{fcfr}  &AAAI & 2021 & RDM  &$l_2$ &S&735.81 &- &- &- \\

DenseLiDAR\cite{gu2021denselidar} &RAL & 2021 & RDM    &$l_2 + l_{grad} + l_{ssim}$ &S&755.41 & - &- &- \\

Zhu et al. \cite{zhu2021robust} &AAAI & 2022  &RDM   &$l_{ud} + l_{ur}$ &S &751.59 & - &PyTorch &- \\
CSPN\cite{cspn}      &ECCV & 2018  &SPM  &$l_2$ &S&1019.64 &17.41 &PyTorch &\checkmark\\
CSPN++\cite{cspn++} &AAAI & 2020 &SPM &$l_2$ &S&743.69 &26 &- &- \\
NLSPN\cite{nlspn}     &ECCV & 2020 &SPM  &$l_1 + l_2$ &S&741.68 &25.84 &PyTorch  &\checkmark\\
PENet\cite{penet}     &ICRA & 2021 &SPM  &$l_2$ &S&730.08 &- &PyTorch &\checkmark\\

DySPN \cite{Lin2022DynamicSP} &AAAI & 2022 &SPM &$l_1 + l_2$ &S&\textbf{709.12} &- &PyTorch &- \\

\hline \hline
SS-S2D (d)\cite{S-d-selfsuper}   &ICRA & 2019 & EFM/EDN   & $l_{photo}$ + $l_{smooth}$ & U &1299.85 &26.1 &PyTorch &\checkmark\\
DFineNet\cite{zhang2019dfinenet}  &Arxiv & 2019 &EFM/EDN  &$l_2$ + $l_{photo}$ + $l_{smooth}$ &S\&U  &943.89 & - &PyTorch &\checkmark \\
DDP\cite{ddp}  &CVPR & 2019 &LFM/DEN  &$l_ 1 + l_{cpn} + l_{photo} + l_{ssim}$ &S\&U &1263.19 &18.8 &TensorFlow  &- \\
DFuseNet\cite{shivakumar2019dfusenet}  &ITSC & 2019 &LFM/DEN &$l_2$ + $l_{stereo}$ + $l_{smooth}$ &S\&U &1206.66 &- &PyTorch &\checkmark \\
VOICED\cite{wong2020unsupervised}    &RAL & 2020  &  LFM/DEN   &$l_1$ + $l_{photo}$ + $l_{smooth}$ &S\&U &1169.97 &9.7 &TensorFlow  &-\\  
AdaFrame\cite{wong2021adaptive} &RAL & 2021  & LFM/DEN &$l_1 + l_{photo} + l_{smooth}$  &S\&U &1125.67 &6.4 &PyTorch &\checkmark  \\
ScaffFusion-S\&U\cite{wong2021scaffnet}     &RAL & 2021  &LFM/DEN  &$l_1 + l_{photo} + l_{smooth} + l_{tp}$ &S\&U &\textbf{847.22} & 7.8 &TensorFlow &\checkmark\\ 
ScaffFusion-U\cite{wong2021scaffnet}     &RAL & 2021  &LFM/DEN  &$l_{photo} + l_{smooth} + l_{tp}$ &U &1121.89  & 7.8 &TensorFlow &\checkmark\\ 
KBNet\cite{wong2021unsupervised}     &ICCV & 2021  & LFM/DEN   &$l_1 + l_{photo} + l_{smooth}$ &S\&U &1069.47 &6.9 &PyTorch &\checkmark \\
Song et al.\cite{song2021self}    &TITS & 2021  & LFM/DEN     &$l_1 + l_{photo} + l_{smooth}$ &S\&U &1216.26 &9.7 &PyTorch &\checkmark \\
\hline
\end{tabular}
\end{center}
\label{table-method-kitti}
\end{table*}


\textbf{VOID \cite{wong2020unsupervised}: } The VOID dataset contains 56 sequences collected with the Intel RealSense D435i camera from both indoor and outdoor scenes, in which 
48 sequences (about 47,000 frames) are designed for training, and the rest of 8 sequences are used for testing.
The resolution of each frame is $640 \times 480$.
Each sequence has three different density levels with 1500, 500, and 150 points.
This dataset was employed to evaluate the methods in \cite{wong2020unsupervised, wong2021adaptive,wong2021scaffnet,wong2021unsupervised,ryu2021scanline}.


\textbf{DenseLivox \cite{DenseLivox}: } 
The DenseLivox dataset is collected with a cheaper Livox LiDAR with much denser depth maps (the density is $88.3\%$) than KITTI. 
DenseLivox also provides some extra data like bound-occlusion and normal. This dataset was employed to evaluate the method in \cite{DenseLivox}.

\subsection{Synthetic Datasets}

\textbf{SYNTHIA \cite{Synthia}: } 
The SYNTHIA is captured in a virtual city that includes street blocks, highways, suburban areas, and other common objects and has 
four different appearances corresponding to four seasons in reality. Different lighting conditions are applied to improve the diversity of virtual RGB images. The dataset has two complementary sets with the image resolution of $960 \times 20$, \textit{SYNTHIA-Rand} and \textit{SYNTHIA-Seqs}. The former (13,400 frames) is obtained randomly within the city, and the latter (200,000 frames) is captured from a virtual vehicle across different seasons. This dataset was employed to evaluate the methods in \cite{max-S-and-D,qu2020depth}.

\textbf{Aerial depth \cite{aerial_depth}: } 
The Aerial depth is a virtual outdoor dataset specially designed for simulating data captured in UVA working conditions. The dataset contains 83797 RGB and depth images from 18 virtual 3D models, and 67435 of them are selected for training and the rest for validation. This dataset was employed to evaluate the method in \cite{aerial_depth}.

\textbf{Virtual KITTI  \cite{gaidon2016virtual}: } This dataset is a virtual version of the KITTI dataset. Five videos of the KITTI (0001/0002/0006/0018/0020) are cloned through the unity engine. The dataset consists of 35 virtual videos (about 17000 frames). Each cloned virtual video is further modified to obtain 7 variations. The modification includes changing features of the objects, the camera's position and orientation, and the lighting condition. This dataset was employed to evaluate the methods in \cite{shivakumar2019dfusenet, qu2020depth, jeon2021abcd, wong2021scaffnet}.

\textbf{SceneNet RGB-D \cite{mccormac2016scenenet}: } 
This dataset contains 5 Million RGBD indoor images from over 15,000 synthetic trajectories with $320\times240$ image resolution. Each trajectory has 300 rendered frames. Due to ray-tracing, 
the generated images can reach the real-photo level quality. This dataset was employed to evaluate the method in \cite{wong2021scaffnet}.

\subsection{Evaluation Metrics}\label{Error-metrics}

Depth completion and monocular depth estimation generally share the same evaluation metrics. We list the most commonly used measures as follows:
\begin{itemize}
    \item \textbf{RMSE}: Root mean squared error defined in equation \eqref{l2_loss}.
    
     \item \textbf{MAE}: Mean absolute error defined in equation \eqref{l1_loss}. 
     \item   \textbf{iRMSE}: RMSE of the inverse depth, defined by $\sqrt{\frac{1}{n}\sum_{i=1}^{n}\left ( \frac{1}{Y_i}-\frac{1}{\hat{Y}_i}\right )^{2}}$.
     \item  \textbf{iMAE}: MAE of the inverse depth, defined by $ \frac{1}{n}\sum_{i=1}^{n}\left | \frac{1}{Y_i}-\frac{1}{\hat{Y}_i}\right |$.
\end{itemize}
The above four measures are metrics commonly used to evaluate models in the KITTI benchmark. Among them, KITTI ranks algorithms in competitions in the order of RMSE. Thus, many previous methods have aimed to choose RMSE ($l_2$) as a loss function to train models.
Besides, several metrics are also frequently used in many methods for depth evaluation, 
such as
\begin{itemize}
    \item \textbf{REL}: Mean relative error defined by $\frac{1}{n}\sum_{i=1}^{n} \frac{| Y_i-\hat{Y}_i |}{\hat{Y}_i} $.
    \item {$\bm \delta$}: Thresholded accuracy defined by  $\max(\frac{Y_i}{\hat{Y}_i},\frac{\hat{Y}_i}{Y_i}) = \delta <\tau$ where  $\tau$ is a given threshold.
\end{itemize}
REL and $\delta$ are commonly used for evaluation of models on indoor datasets, e.g., NYU-v2.

Evaluation of depth maps is an open issue. The above metrics cannot precisely measure the quality of reconstructed compositional patterns such as objects. Therefore, researchers also attempted to propose new evaluation metrics. In \cite{Hu2019RevisitingSI}, object boundaries extracted from the depth map are measured. Koch et al. \cite{Koch2020ComparisonOM} introduced the planarity error and location accuracy of depth boundaries.
Jiang et al. \cite{jiang2021plnet} proposed two metrics for quantifying the flatness of planes and the straightness of lines for depth maps. However, owing to the lack of dense ground truth, such metrics are still difficult to be applied to depth completion.

\section{Experimental Analyses}
\label{result}
In this section, we compare and review previous methods from comprehensive aspects. Specifically, we select some representative works from each category and elucidate their major characteristics, including network structure, loss function, learning strategy, model performance, etc.
Table~\ref{table-method} and Table~\ref{table-method-kitti} show a comparison of existing unguided and RGB guided methods on the KITTI dataset, respectively, where the RMSE values are taken from either the public KITTI benchmark or the original papers.
Table~\ref{table-method-nyuv2} shows a comparison of RGB guided methods on the NYU-v2 dataset.
Besides, Table~\ref{table-method-void} shows a comparison of relevant methods on the VOID dataset. Note that we use \textbf{S}, \textbf{U}, and \textbf{S\&U} to denote supervised methods, purely unsupervised methods without applying depth consistency, and unsupervised methods with depth consistency on only valid depth points from the sparse inputs in Table~\ref{table-method}, ~\ref{table-method-kitti}, ~\ref{table-method-nyuv2}, and ~\ref{table-method-void}. 
We use the RMSE metric for performance comparison.
In the following sections, our findings are summarized.

\begin{table*}[t]
\caption{Summary of essential characteristics of existing RGB guided methods on the NYU-v2 dataset. For denoting loss functions, we omit the coefficient of each loss term for simplicity. S and U denotes supervised learning and unsupervised learning of models, respectively.}
\renewcommand\arraystretch{1.2}
\scriptsize
\begin{center}
\begin{tabular}
{|m{0.110\textwidth}<{\raggedleft}m{0.05\textwidth}<{\centering}m{0.05\textwidth}<{\centering}|m{0.1\textwidth}<{\centering}m{0.16\textwidth}<{\centering}m{0.05\textwidth}<{\centering}|m{0.05\textwidth}<{\centering}m{0.05\textwidth}<{\centering}m{0.08\textwidth}<{\centering}m{0.05\textwidth}<{\centering}|}
\hline
\textbf{Method}  &\textbf{Publication}  &\textbf{Year} & \textbf{Type} & \textbf{Loss Function} &\textbf{Learning} & \textbf{RMSE (mm)} &\textbf{Params (M)}  &\textbf{Platform} & \textbf{Code} \\ 
\hline
3coef\cite{depth_coefficient}&CVPR &2019 &EFM/EDN  &$l_{ce}$ &S &131 &- & TensorFlow & \checkmark \\
Long et al.\cite{long2021depth} &JVCIR   & 2021 &EFM/EDN  &$l_1 + l_{cosine}$ &S &100 &- &- &- \\
DFuseNet\cite{shivakumar2019dfusenet} &ITSC & 2019 &LFM/EDN &$l_2$ + $l_{stereo}$ + $l_{smooth}$ &S\&U &219 &- &PyTorch &\checkmark \\
MS-Net[LF]\cite{eldesokey2019confidence}  &TPAMI & 2019 &LFM/DEN  &$l_{huber}$ &S &129 &0.356 &PyTorch &\checkmark \\

KBNet\cite{wong2021unsupervised}   &ICCV  & 2021  & LFM/DEN &$l_1 + l_{photo} + l_{smooth}$  &S\&U &105 &6.9 &PyTorch &\checkmark \\
SelfDeco\cite{choi2021selfdeco} &ICRA  & 2021  &LFM/DEN &$l_1 + l_{photo} + l_{smooth}$  &S\&U &178 & &PyTorch &- \\

GuideNet\cite{learning-guided} &TIP & 2020 &LFM/DEDN &$l_2$ &S &101 &62.62 &PyTorch &\checkmark\\ 

RigNet\cite{yan2021rignet}  &Arxiv  & 2021  & LFM/DEDN  &$l_2$ &S &\textbf{90} &- &PyTorch &- \\
PwP\cite{xu2019depth}    &ICCV   & 2019  & E3DR/ISNR &$l_2 + l_{normal}$ &S &112 &28.99 &PyTorch  &28.99 \\
DeepLiDAR\cite{DeepLiDAR} &CVPR & 2019 &E3DR/ISNR &$l_2 + l_{normal}$ &S &115 &144 &PyTorch  &\checkmark\\
ACMNet\cite{zhao2021adaptive} &TIP   & 2021 &E3DR/3DAC  &$l_2$ &S &105 &1.35 &PyTorch &\checkmark \\
FCFR-Net\cite{fcfr} &AAAI  & 2020 &RDM      &$l_2$ &S &106 &- &- &- \\
KernelNet \cite{liu2021learning}  &TIP &2021 &RDM  &$l_1 + l_{ce} + l_{grad}$ &S &111 &- &PyTorch &\checkmark \\
CSPN\cite{cspn}    &ECCV  & 2018  &SPM &$l_2$ &S &117 &17.41 &PyTorch &\checkmark\\
CSPN++\cite{cspn++} &AAAI & 2020 &SPM &$l_2$ &S &115 &26 &- &- \\
NLSPN\cite{nlspn}   &ECCV  & 2020 &SPM     &$l_1$  &S &92 &25.84 &PyTorch  &\checkmark\\
DySPN\cite{Lin2022DynamicSP}   &AAAI  & 2022 &SPM    &$l_1$ + $l_2$ &S &\textbf{90} &- &PyTorch  &-\\

\hline
\end{tabular}
\end{center}
\label{table-method-nyuv2}
\end{table*}

\subsection{Main Characteristics of Existing Methods}

\begin{enumerate}
    \item  A relatively smaller number of prior works employ the route of performing completion from the sparse depth input. In comparison, more recent works are RGB-guided, among which the majority route is to perform late fusion of RGB and depth images instead of early fusion.
    
    \item PyTorch is the most popular deep learning library for implementing depth completion methods. The overwhelming majority of previous studies implement their methods with PyTorch.
    
    \item KITTI is the most popularly used evaluation benchmark. Almost all leading methods provide results on this dataset. Moreover, NYU-v2 is the second most popular dataset. Since depth maps of NYU-v2 are captured by Kinect, previous works implement their methods by randomly and uniformly sampling 200 or 500 pixels as valid depth points.  Besides, VOID is also a frequently used benchmark for recent unsupervised methods. 
    
    \item More complicated neural network modules have been recently developed to advance the performance of depth completion models. For example, many methods propose to embed surface normal, affinity matrices, and residual maps into their network models. 
    
    \item  The learning objectives identified for depth completion tasks are intuitive and relatively straightforward to optimize. For example,  many methods penalize just $l_1$ or $l_2$ loss of depth maps, and still achieve good performance.

\end{enumerate}

\subsection{Unguided and Guided Methods}
There are two benefits of unguided methods. First,
 unguided methods are more robust to environments with light or weather changes since they only take sparse depth maps as inputs. Moreover, for the same reason, they are more computationally efficient. 
 However, unguided methods show inferior performance due to the lack of semantic cues and the irregular distribution of captured depth points.
As seen in Table~\ref{table-method}, the best unguided method \cite{from_depth_what} yields RMSE of 901.43 millimeters on the KITTI dataset. Note that \cite{from_depth_what} also uses RGB images to guide model training. The best result obtained using an RGB-free method in both the training and inference stage is demonstrated in \cite{huang2019hms} with RMSE of 937.48. 
On the other hand, as seen in Table~\ref{table-method-kitti}, the best RGB guided method, i.e., DySPN, demonstrates a significantly better result with RMSE of 709.12. Moreover, many RGB guided methods can easily beat the best unguided approach. Specifically, except for  3coef \cite{depth_coefficient}, EncDec-Net[EF]  \cite{eldesokey2019confidence}, Morph-Net \cite{dimitrievski2018learningmorph} and CSPN \cite{cspn}, all other RGB guided methods with  supervised learning outperform HMS-Net, showing the advance of leveraging RGB information.
Another difference is that unguided methods cannot utilize additional unsupervised losses derived from images, e.g., photometric loss.

\subsection{Comparison of RGB Guided Methods}

For RGB guided methods, from Table~\ref{table-method-kitti}, we can observe the following results:
\begin{itemize}
    \item  Early fusion models generally underperform other types of methods.

\item For later fusion approaches, although a considerable number of methods are built on DEN, approaches \cite{learning-guided,yan2021rignet} based on DEDN demonstrate more significant performance improvement.

\item Explicit 3D representation methods, SPN-based methods, and residual depth methods show more advanced performance and generally outperform other approaches.
\end{itemize}

More specifically, the Top-10 performing methods on the KITTI dataset
are (i) four SPN-based models; DySPN \cite{Lin2022DynamicSP}, PENet \cite{penet}, NLSPN \cite{nlspn}, and CSPN++ \cite{cspn++}, (ii) two residual depth models; FCFR-Net \cite{fcfr} and \cite{zhu2021robust}, (iii) two late fusion methods built on DEDN; RigNet \cite{yan2021rignet} and GuideNet \cite{learning-guided}, and (iv) two explicit 3D representation models; ACMNet \cite{zhao2021adaptive} and 2D-3D FuseNet \cite{2d-3d}. 
Based on that, we can say that the naive fusion strategy such as aggregating inputs at an early stage or concatenating features extracted by a dual-encoder network in late stage is not sufficient for achieving satisfactory performance. 
The common feature of the Top-10 performing methods is that they propose to either explicitly model geometric relationship of depth points by applying 3D-aware convolution as ACMNet and 2D-3D FuseNet, refinement with residual depth map as residual depth models and affinity matrix as SPN-based methods; or learn more effective guided kernel to weigh depth features with a complicated network design as RigNet and GuideNet.

\begin{table*}[t]
\caption{Summary of essential characteristics of existing RGB guided methods on the VOID dataset. For denoting loss functions, we omit the coefficient of each loss term for simplicity. S and U denotes supervised learning and unsupervised learning of models, respectively. $^{*}$ and $^{**}$ denote results taken from \cite{wong2021unsupervised} and {paperswithcode \protect \footnotemark}, respectively.}
\renewcommand\arraystretch{1.2}
\scriptsize
\begin{center}
\begin{tabular}
{|m{0.136\textwidth}<{\raggedleft}m{0.05\textwidth}<{\centering}m{0.05\textwidth}<{\centering}|m{0.1\textwidth}<{\centering}m{0.16\textwidth}<{\centering}m{0.05\textwidth}<{\centering}|m{0.05\textwidth}<{\centering}m{0.05\textwidth}<{\centering}m{0.08\textwidth}<{\centering}m{0.03\textwidth}<{\centering}|}
\hline
\textbf{Method}  &\textbf{Publication}  &\textbf{Year} & \textbf{Type} & \textbf{Loss Function} &\textbf{Learning} & \textbf{RMSE (mm)} &\textbf{Params (M)}  &\textbf{Platform} & \textbf{Code} \\ 
\hline
SS-S2D$^{*}$ \cite{S-d-selfsuper}   &ICRA & 2019 & EFM/EDN   & $l_1 +l_{photo}$ + $l_{smooth}$ & S\&U &243.84 &27.8 &- &- \\
DDP$^{*}$ \cite{ddp}  &CVPR & 2019 &LFM/DEN  &$l_ 1 + l_{cpn} + l_{photo} + l_{ssim}$ &S\&U &222.36 &18.8 &-  &- \\
NLSPN$^{**}$ \cite{nlspn}  &ECCV & 2020 &SPN  &$l_ 1 + l_2$ &S &79.12 &25.84 &-  &- \\
VOICED\cite{wong2020unsupervised}    &RAL & 2020  &  LFM/DEN   &$l_1$ + $l_{photo}$ + $l_{smooth}$ &S\&U & 146.40 &9.7 &TensorFlow  &-\\  
Ryu et al.\cite{ryu2021scanline} &RAL &2021 &LFM/DEN & $l_2 + l_{smooth}$ &S  &181.42 &-  &- &-  \\
AdaFrame\cite{wong2021adaptive} &RAL & 2021  & LFM/DEN &$l_1 + l_{photo} + l_{smooth}$  &S\&U &135.93 &6.4 &PyTorch &\checkmark  \\
ScaffFusion\cite{wong2021scaffnet}     &RAL & 2021  &LFM/DEN  &$l_1 + l_{photo} + l_{smooth} + l_{tp}$ &S\&U &119.14 &7.8 &TensorFlow &\checkmark\\ 
KBNet\cite{wong2021unsupervised}     &ICCV & 2021  & LFM/DEN   &$l_1 + l_{photo} + l_{smooth}$ &S\&U &95.86 &6.9 &PyTorch &\checkmark \\
\hline
\end{tabular}
\end{center}
\label{table-method-void}
\end{table*}

 Consistent results are also observed in analyses on the NYU-v2 dataset. As shown in Table~\ref{table-method-nyuv2}, the best results are demonstrated by DySPN and RigNet. Besides, GuideNet, ACMNet, FCFR-net, and NLSPN also show improved performance compared to other methods. 

Intuitively, the performance of depth completion has the potential to be further improved by aggregating core technical components of the above methods. For instance, by taking advantage of 3D representation networks and spatial propagation networks, we can not only learn the 3D relationship within the model in a feature space but also apply post-refinement with an affinity matrix in output space. In addition, we can also incorporate a DEDN with guided kernel learning into residual depth learning models. Such combinations are straightforward, nevertheless, can be considered in practical applications to pursue high accuracy.

\subsection{Results of unsupervised Approaches}
\label{results_of_unsupervised_learning}

The bottom of Table~\ref{table-method-kitti} shows methods with unsupervised photometric loss. 
Results of purely unsupervised methods (without using depth consistency loss) are calculated by aligning the scale of the predicted depth map to the scale of ground truth. 
First, for methods without leveraging depth consistency, such as SS-S2D (d) \cite{S-d-selfsuper} and ScaffFusion-U \cite{wong2021scaffnet}, we can see that purely unsupervised methods demonstrate unsatisfactory performance. Second, we also observe that their performances are still inferior to supervised methods even leveraging both depth consistency loss and additional photometric loss. As also discussed in Sec.~\ref{real-world datasets}, this is because these methods \cite{wong2021scaffnet,wong2021unsupervised,ddp,song2021self,choi2021selfdeco} use sparser depth maps as ground truths with a density of $5\%$ than supervised methods with a density of $30\%$.
Among these methods, ScaffFusion \cite{wong2021scaffnet}, DFineNet \cite{zhang2019dfinenet}, and KBNet \cite{wong2021unsupervised} demonstrate better performance than other approaches.
Similar results are also observed in Table~\ref{table-method-nyuv2} where supervised approaches outperformed unsupervised methods. This is not surprising since supervised methods could use pixel-wise ground truth depth maps for training on the NYU-v2 dataset.

 Table~\ref{table-method-void} provides results for several approaches evaluated on the VOID dataset. 
As observed, the performance has been continuously improved from the early VOICED \cite{wong2020unsupervised} to the recent KBNet \cite{wong2021unsupervised}.
Most works of Wong et al. apply pre-densification to sparse depth inputs, such as  employing a learning based spatial pyramid pooling (SPP) block in \cite{wong2021scaffnet}. As argued in \cite{wong2021unsupervised}, the max-pool layers in the SPP block tend to lose details in close range. Therefore, in KBNet, both max-pooling and min-pooling are implemented to ensure that the network can extract more comprehensive depth features. We believe this plays an important role in improving the KBNet's accuracy.

Overall, KBNet \cite{wong2021unsupervised} and ScaffFusion \cite{wong2021scaffnet} achieved the first and the second highest accuracy among unsupervised approaches on the VOID dataset. However, they still underperform the supervised NLSPN. It reveals that the photometric loss used by current unsupervised approaches is still not fully reliable and accurate as they are vulnerable to outliers, e.g., dynamic objects, sky, and transparent objects, that are ubiquitous in real-world scenarios.

\footnotetext{https://paperswithcode.com/sota/depth-completion-on-void}

\section{Open Challenges and Future Directions}
\label{discussion}

\subsection{Depth Mixing Problem}

The depth mixing problem, also called the depth smearing problem, is attributed to the difficulty of correctly identifying pixels near object boundaries, and usually causes blurry edges and artifacts.
In order to alleviate this problem, 3coef. \cite{depth_coefficient} formulates depth completion as a one-hot encoding problem by dividing a depth map into a set of bins with fixed depth ranges. Imran et al. \cite{Imran2021DepthCW} isolate the foreground and background depths in occlusion-boundary regions and models them, respectively.
NLSPN \cite{nlspn} makes the network learn non-local relative neighbors such that the pixels can be separated during an iterative propagation. A more simple way of achieving this separation process is to leverage the K-nearest algorithm \cite{2d-3d,zhao2021adaptive,dan-conv}. Besides, a boundary consistency network was added after depth completion to encourage predicting more sharp and clear boundaries \cite{huang2019indoor,tao2021dilated}. However, this problem is still difficult for depth estimation tasks and needs to be continuously investigated.

\subsection{Flawed Ground Truth}\label{flawef ground truth}

Another problem is the existence of defects in ground truth depths. First, unlike semantic segmentation, none of the existing real-world datasets can provide pixel-wise ground truth because of the limitation of depth sensors. Although many existing methods are trained in a supervised way, most pixels cannot be sufficiently supervised. Second, the semi-dense annotations are not entirely reliable due to outliers caused by occlusions, dynamic objects, etc. To overcome the sparsity problem, some researchers \cite{S-d-selfsuper,song2021self} turn to self-supervised frameworks to alleviate the lack of ground truth depths. To cope with the second problem, Zhu et al. \cite{zhu2021robust} handle outliers by incorporating uncertainty estimation into the depth completion network.
Besides, a few works \cite{to_complete_or_to_estimate,multitask_gan} leverage synthetic datasets for model training. However, the domain gap between real-world and synthetic data prevents a wide application of these methods. 
Despite the above efforts made by previous studies, it is still an open issue how to exclude the effects of unreliable depths, and there are still lots of room for improvement.

\subsection{Lightweight Networks}
Most previous methods have complex network structures with a large number of parameters. 
Moreover, many of them take two-stage coarse-to-refinement prediction.
Thus, these methods are time-consuming and require high usage of hardware resources.
However, for applications such as autonomous driving and robotic navigation, computation resources are limited and real-time inference is required. 
Although a few prior studies \cite{tao2021dilated,dan-conv,eldesokey2019confidence,bai2020depthnet} have partially considered the real-time inference problem, they suffer from inferior performance. Besides, the network design is essentially empirical. 
Following advances in monocular depth estimation, we can further apply several techniques, such as applying knowledge distillation \cite{hu2021boosting}, network compression \cite{Wofk2019FastDepthFM}, and neural architecture search \cite{huynh2022lightweight}.
Without sacrificing too much accuracy, developing lightweight methods with fast inference speed has enormous potential for real-world deployment, thus, is a valuable and practical research point in future work. 

\subsection{Un-/self-supervised Frameworks}
As discussed before, un-/self-supervised learning frameworks are solutions commonly employed in the absence of dense ground truths. 
 As discussed in Sec.~\ref{results_of_unsupervised_learning}, the accuracy of current un-/self-supervised methods is still lower compared to supervised methods since they apply depth consistency to only valid depth points from sparse inputs and cannot leverage ground truth depth points as many as used by supervised methods. On the other hand, the photometric loss will only be effective when the predicted depth maps are close enough to the ground truth. However, that is still challenging due to the fact that the photometric loss is particularly susceptible to noises, moving objects, texture-less regions.
 Thus, there is much room for further improvement of unsupervised methods. Since this kind of methods is not robust to dynamic objects, distant regions, etc., the improvements can be brought by leveraging more effective
network structures for performing auxiliary tasks, such as pose estimation and outlier removal.

\subsection{Loss Functions and Evaluation Metrics}
Employment of proper loss functions is also critical to achieving satisfactory performance for depth completion. Commonly used loss functions are usually defined by a weighted sum of $l_{2}$  or $l_{1}$ loss functions with other auxiliary loss functions, e.g., smoothness loss and SSIM loss. However, as discussed in \cite{depth_coefficient}, both $l_{1}$  and $l_{2}$ loss functions have their own drawbacks. The choice of them is usually dataset dependent. Similarly, current metrics cannot precisely measure the quality of scene structures.
Although several new metrics have been introduced in  \cite{Hu2019RevisitingSI,depth_coefficient,Koch2020ComparisonOM,jiang2021plnet} for evaluating depth maps, they have not gained broad popularity. 
Thus, designing more effective loss functions and convincing evaluation metrics is also a potential future research direction.

\subsection{Domain Adaptation}
Current benchmark datasets face the challenge of the lack of reliable depth points. Moreover,
the data is captured under ideal lighting conditions in limited scenarios. Thus, models trained using this type of data have no guarantee of generalization in different working conditions and domains.
Accordingly,
it is reasonable to manipulate deep networks in simulated environments. Thereby, we can have not only per-pixel ground truth but also changeable lighting or weather conditions with a great number of different scenarios. Moreover, it encourages the development of more advanced methods that are difficult to be implemented in the real world. 
The challenge is then how to transfer the model from simulated environments to real-world scenarios.
A few works explored domain adaptation methods for depth completion \cite{lopez2020project, to_complete_or_to_estimate}. However, this under-explored problem remains unknown and is worthy of further exploration.

\subsection{Transformer-based Network Structures}
Recently, visual transformers (ViT) have attracted extensive attention and continuously introduced new state-of-the-art results for many perception tasks, including classification \cite{DosovitskiyB0WZ21}, semantic segmentation \cite{Strudel2021SegmenterTF}, object detection \cite{zhang2021vit} and monocular depth estimation \cite{Bhat2021AdaBinsDE}. Unlike CNNs, ViT receives a set of image patches as input and uses self-attention for local and global feature interactions. It may bring a new paradigm shift for depth completion where more effective multi-modality data fusion and novel strategies for handling input sparsity may exist.

\subsection{Visualization and Interpretability}
A few works have attempted to understand and visualize the mechanism of CNNs for monocular depth estimation. It is shown in \cite{hu2019analysis,hu2019visualization,dijk2019neural} that CNNs tend to use some monocular cues from RGB images for inferring depths. In addition, as observed in \cite{you2021towards} that the features generated inside CNNs are highly disentangled and activated to different depth ranges. An intriguing question is what will be different if we estimate depths when a few sparse depth points are available in inputs. Exploring and answering the above question is essential to the interpretation of learning based approaches, and has promising applications for improvement of their generalization ability, e.g., facilitating domain adaptation; and robustness of deep learning based depth completion methods.

\subsection{Robustness to Different Sensors}
Existing methods are only applicable to particular sensors. For instance, the most frequently used KITTI dataset is captured by a 64-line LiDAR. There is no guarantee that previous methods can be applied to lower scanline sensors, such as 32-line, 16-line LiDARs, and 1-line LiDARs. As demonstrated by \cite{lu2021sgtbn,S-d-selfsuper,Yoon2020BalancedDC,ryu2021scanline}, the performance degradation is significant from a 64-line sensor to lower scanline sensors. Hence, maintaining the same level of accuracy for lower scanline sensors is challenging.
This under-explored problem is also practical in real-world applications since higher scanline sensors are more expensive than lower ones.
Therefore, ensuring the accuracy of learning based methods for various lower scanline sensors is also an important and valuable research topic.

\section{Conclusion}\label{conclusion}
In this article, we present a comprehensive survey of deep learning based depth completion methods. Our review covers traditional and state-of-the-art
network structures, loss functions, learning strategies, benchmark datasets, and evaluation metrics. 
To depict the evolution process and draw the connections between existing works, we provide a fine-grained taxonomy that categorizes existing methods by jointly considering network structures and main technical contributions. Moreover, we visualize the main characteristics of existing methods as well as their quantitative performance on the most popular benchmark datasets to provide an intuitive and straightforward comparison. We then perform in-depth analyses that summarize their performances, similarities, and differences. Finally, we provide open challenges and promising future research directions. 
Through the above efforts, we hope our work can 
help readers navigate this field.

\bibliographystyle{IEEEtranS}
\bibliography{egbib}

\begin{IEEEbiography}[{\includegraphics[width=1in,height=1.25in,clip,keepaspectratio]{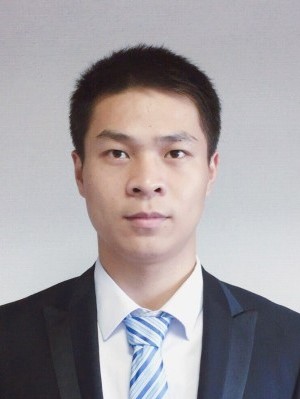}}]{Junjie Hu} (Member, IEEE) received the M.S. and Ph.D. degrees from the Graduate School of Information Science, Tohoku University, Sendai, Japan, in 2017 and 2020, respectively. He is currently a Research Scientist with the Shenzhen Institute of Artificial Intelligence and Robotics for Society. His research interests include  machine learning, computer vision, and robotics.
\end{IEEEbiography}

\begin{IEEEbiography}[{\includegraphics[width=1in,height=1.25in,clip,keepaspectratio]{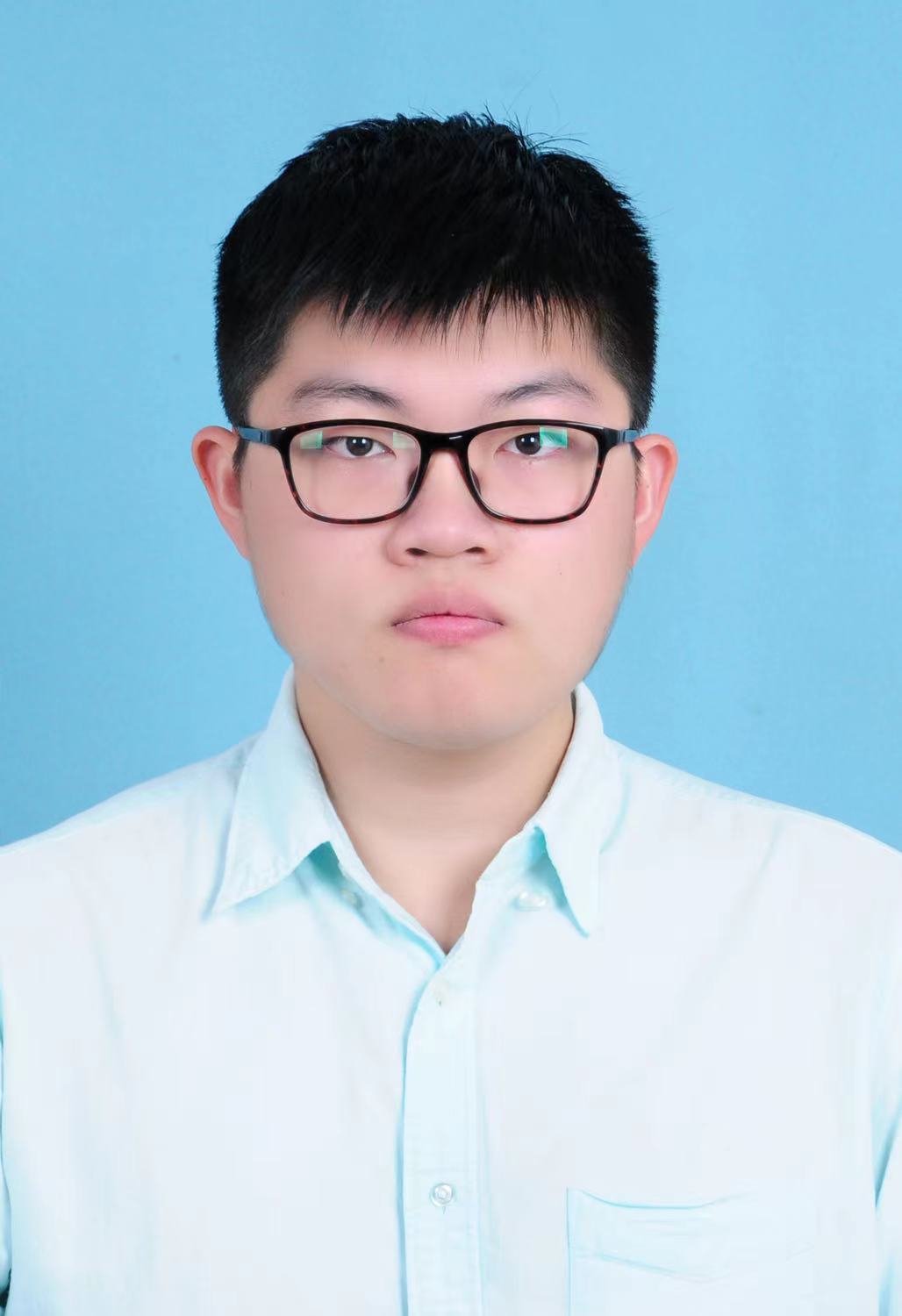}}]{Chenyu Bao} received the B.Eng degree from the School of Mechanical Science and Engineering, HuaZhong University of Science and Technology in 2022. He is currently a graduate student in the School of Science and Engineering, Chinese University of Hong Kong, Shenzhen. His research interests include computer vision and robot perception.
\end{IEEEbiography}

\begin{IEEEbiography}[{\includegraphics[width=1in,height=1.25in,clip,keepaspectratio]{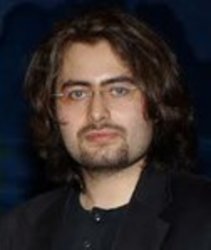}}]{Mete Ozay}  (M’09) received the B.Sc., M.Sc., Ph.D. degrees in mathematical physics, information systems, and computer engineering \& science from METU, Turkey. He has been a visiting Ph.D. and fellow in the Princeton University, USA, a research fellow in the University of Birmingham, UK, and an Assistant Professor in the Tohoku University, Japan. His current research interests include pure and applied mathematics,  theoretical computer science \& neuroscience.
\end{IEEEbiography}

\begin{IEEEbiography}[{\includegraphics[width=1in,height=1.25in,clip,keepaspectratio]{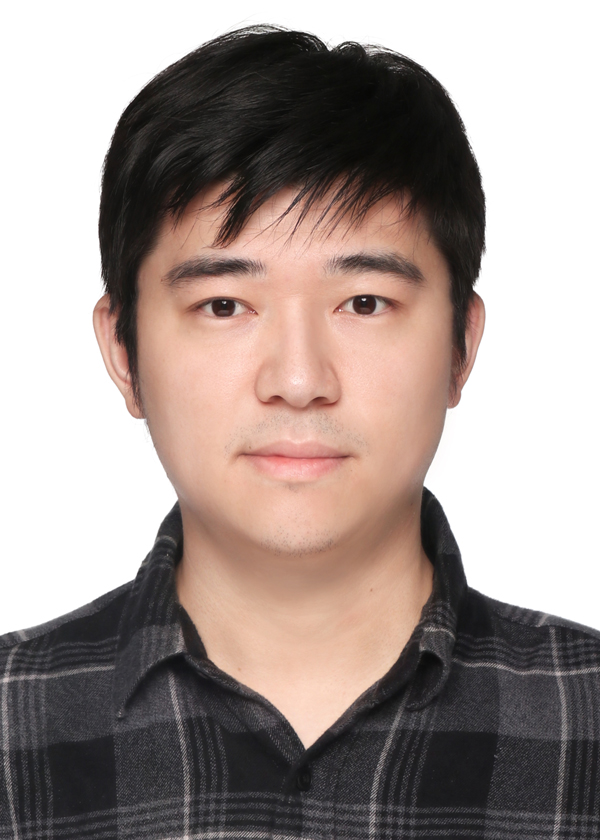}}]{Chenyou Fan} serves as Associate Profession of the School of Artificial Intelligence, South China Normal University, China. 
He received the B.S. degree in computer science from the Nanjing University, China, in 2011, and the M.S. and Ph.D. degrees from Indiana University, USA, in 2014 and 2019, respectively.   His research interests include machine learning and computer vision.
He served in the program committee of CVPR,
NeurIPS, ACM MM and top AI journals for more
than 20 times.
\end{IEEEbiography}

\begin{IEEEbiography}[{\includegraphics[width=1in,height=1.25in,clip,keepaspectratio]{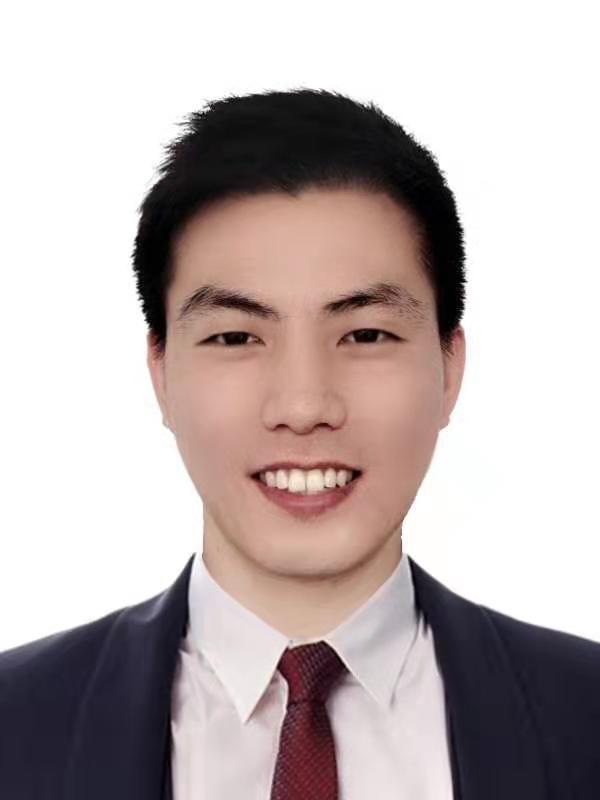}}]{Qing Gao} received his Ph.D. degree in the State Key Laboratory of Robotics, Shenyang Institute of Automation (SIA),
Chinese Academy of Sciences (CAS), Shenyang, China, in 2020. He is currently a Research Scientist with the Shenzhen Institute of Artificial Intelligence and Robotics for Society. His research interests include robotics, artificial intelligence, machine vision and human–robot interaction.
\end{IEEEbiography}

\begin{IEEEbiography}[{\includegraphics[width=1in,height=1.25in,clip,keepaspectratio]{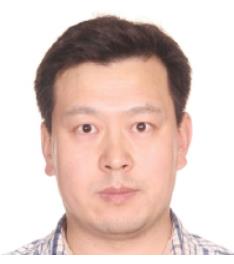}}]{Honghai Liu} (Fellow, IEEE) received the Ph.D. degree in intelligent robotics from King’s College London, London, U.K., in 2003.
He is a Professor with the Harbin Institute of Technology (Shenzhen), Shenzhen, China. He is also a Chair Professor of Human–Machine Systems with the University of Portsmouth, Portsmouth, U.K. His research interests include multi-sensory data fusion, pattern recognition, intelligent video analytics, intelligent robotics, and their practical applications. He is an Associate Editor of the  IEEE Transactions on Industrial Electronics, the IEEE Transactions on Industrial Informatics, and the IEEE Transactions on Cybernetics. 
\end{IEEEbiography}

\begin{IEEEbiography}[{\includegraphics[width=1in,height=1.25in,clip,keepaspectratio]{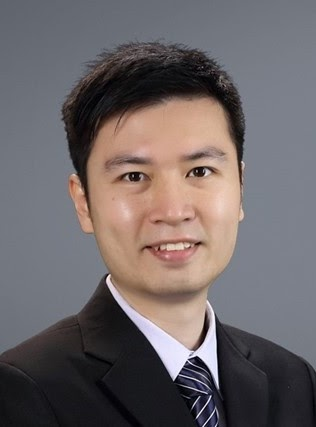}}]{Tin Lun Lam} (Senior Member, IEEE)  received the B.Eng. (First Class Hons.) and Ph.D. degrees in robotics and automation from the Chinese University of Hong Kong, Hong Kong, in 2006 and 2010, respectively. 
He is currently an Assistant Professor with the Chinese University of Hong Kong, Shenzhen, China, and the Director of Center for the Intelligent Robots, Shenzhen Institute of Artificial Intelligence and Robotics for Society. He has authored or coauthored two monographs and more than 50 research papers in top-tier international journals and conference proceedings in robotics and AI. His research interests include multirobot systems, field robotics, and collaborative robotics.
\end{IEEEbiography}

\end{document}